\pdfoutput=1

\documentclass[11pt]{article}

\usepackage[preprint]{acl}

\usepackage{times}
\usepackage{latexsym}

\usepackage[T1]{fontenc}

\usepackage[utf8]{inputenc}

\usepackage{microtype}

\usepackage{inconsolata}

\usepackage{graphicx}
\usepackage{amsthm,amsmath,amssymb}
\usepackage{mathrsfs}
\usepackage{booktabs}
\usepackage{algorithm}
\usepackage{algpseudocode}
\usepackage{comment}
\usepackage{multirow}
\usepackage{bbding}
\usepackage{array}
\usepackage{tcolorbox}
\usepackage{colortbl}
\usepackage{xcolor}
\usepackage{enumitem}
\usepackage{threeparttable}
\usepackage{CJKutf8}
\usepackage{minitoc}
\usepackage{hyperref}
\usepackage{titletoc}
\usepackage[page,header]{appendix}

\newcommand{\blanksymbolfootnote}[1]{%
  \renewcommand{\thefootnote}{}
  \footnote{#1}%
  \setcounter{footnote}{0} 
  \renewcommand{\thefootnote}{\arabic{footnote}}
}

\newcommand{\scriptnumber}[1]{\textcolor{gray}{\small (#1)}}

\title{\textsc{HPSS}: Heuristic Prompting Strategy Search for LLM Evaluators}

\author{Bosi Wen$^{1,\dagger}$ \quad Pei Ke$^{2}$\quad Yufei Sun$^{4}$\quad Cunxiang Wang$^{3,4}$\quad Xiaotao Gu$^{3}$ \\
\textbf{Jinfeng Zhou$^{1}$\quad  Jie Tang$^{4}$\quad Hongning Wang$^{1}$ \quad Minlie Huang$^{1,\ddagger}$} \\
$^1$The Conversational Artificial Intelligence (CoAI) Group, Tsinghua University \\
$^2$University of Electronic Science and Technology of China \quad $^3$Zhipu AI \\
$^4$The Knowledge Engineering Group (KEG), Tsinghua University \\
\tt\small wbs23@mails.tsinghua.edu.cn, aihuang@tsinghua.edu.cn \\}

\begin{document}
\maketitle

\blanksymbolfootnote{$^\dagger$Work done when this author interned at Zhipu AI.}
\blanksymbolfootnote{$^\ddagger$Corresponding author}

\vspace{-6mm}
\begin{abstract}
Since the adoption of large language models (LLMs) for text evaluation has become increasingly prevalent in the field of natural language processing (NLP), a series of existing works attempt to optimize the prompts for LLM evaluators to improve their alignment with human judgment. 
However, their efforts are limited to optimizing individual factors of evaluation prompts, such as evaluation criteria or output formats, neglecting the combinatorial impact of multiple factors, which leads to insufficient optimization of the evaluation pipeline. 
Nevertheless, identifying well-behaved prompting strategies for adjusting multiple factors requires extensive enumeration.
To this end, we comprehensively integrate 8 key factors for evaluation prompts and propose a novel automatic prompting strategy optimization method called Heuristic Prompting Strategy Search (HPSS). 
Inspired by the genetic algorithm, HPSS conducts an iterative search to find well-behaved prompting strategies for LLM evaluators. 
A heuristic function is employed to guide the search process, enhancing the performance of our algorithm.
Extensive experiments across four evaluation tasks demonstrate the effectiveness of HPSS, consistently outperforming both human-designed evaluation prompts and existing automatic prompt optimization methods. 
Our code is available at \url{https://github.com/thu-coai/HPSS}.
\end{abstract}

\section{Introduction}

Evaluation is a long-standing critical and challenging task in NLP \cite{celikyilmaz2020evaluation, chang2024survey}.
Recently, with the advent of advanced large language models (LLMs) such as GPT-4, LLM-based evaluators have been widely used for various natural language generation tasks \cite{wang-etal-2023-chatgpt, liu-etal-2023-g, zheng2023judging, hu-etal-2024-llm, ke-etal-2024-critiquellm}. 
Leveraging the LLMs' strong language understanding and instruction-following capabilities, the evaluation task can be addressed via an evaluation prompt that includes a task description, evaluation rules, and the text to be assessed
\cite{chen-etal-2023-exploring-use, ke-etal-2023-decompeval}. 
Various studies have reported that LLMs can provide accurate and interpretable evaluation results of text quality
on the specified aspect, offering stronger generalization ability compared with previous evaluation methods \cite{10.3115/1073083.1073135, lin-2004-rouge, bert-score, zhong-etal-2022-towards}.

\begin{figure}[t]
\scriptsize
    \centering
    \includegraphics[width=1.0\linewidth]{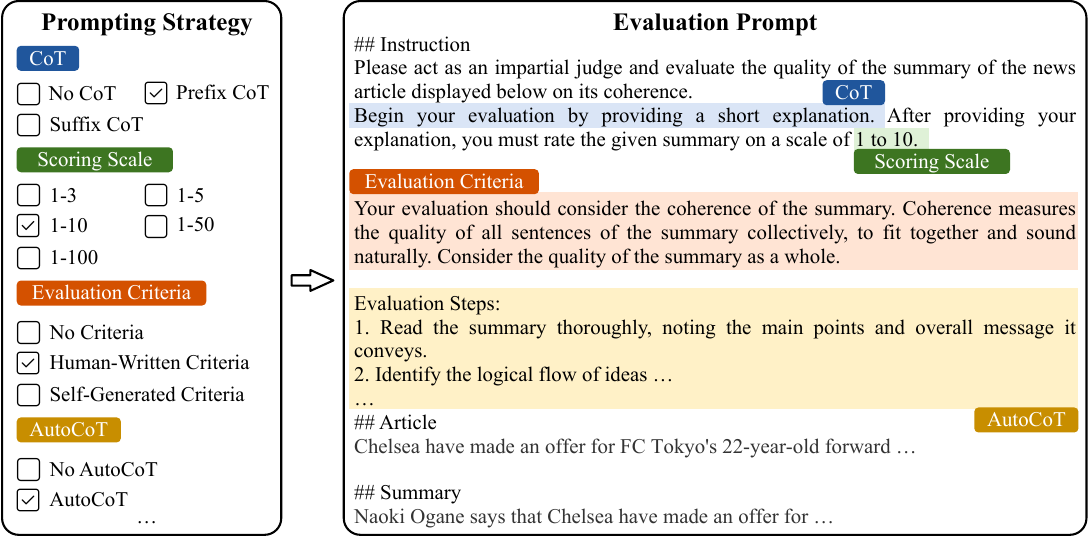}
    \vspace{-4mm}
    \caption{An example of prompting strategy and its corresponding evaluation prompt for assessing the coherence of text summarization, where some factors are highlighted and others are not shown.}
    \vspace{-6mm}
    \label{fig:example}
\end{figure}

Despite the advantages mentioned above, LLM-based evaluation methods are still imperfect in aligning with human judgments \cite{wang-etal-2023-chatgpt, liu2024aligning}, which cannot fully serve as an alternative to human evaluation as the gold standard. 
An emerging line of research attempts to optimize 
evaluation prompts for LLM evaluators to improve their alignment with human judgments from multiple aspects, including optimizing evaluation criteria \cite{liu-etal-2024-calibrating, liu-etal-2024-hd}, modifying output formats \cite{chiang-lee-2023-closer, chu2024better}, and providing in-context examples \cite{kim-etal-2023-better, jain-etal-2023-multi, huang2024empirical}.

However, we argue that these methods are only limited to optimizing a single factor of evaluation prompts (e.g., evaluation criteria or output formats) but neglect the prompting strategy for adjusting multiple factors \citep{kim-etal-2023-better}. 
Given that the evaluation prompts comprise multiple components \cite{gao2024llm}, each encompassing various factors that simultaneously affect the performance of LLM evaluators, it is necessary to collectively adjust these factors to fully harness its evaluation capability. 

Nevertheless, due to the vast state space involved in making these adjustments, extensive enumeration is required to identify well-behaved prompting strategies, highlighting the need for effective automatic optimization methods.

To this end, we comprehensively integrate a series of prompting strategies for LLM evaluators from previous works, select 8 key factors, 
and propose the \textbf{\textit{H}}euristic \textbf{\textit{P}}rompting \textbf{\textit{S}}trategy \textbf{\textit{S}}earch (HPSS) algorithm to automatically optimize prompting strategies for LLM evaluators tailored to specific evaluation aspects.
Inspired by the genetic algorithm \cite{mitchell1980need}, 
HPSS gradually mutates the previous prompting strategies and selects candidate strategies based on their performance on a validation dataset. 
Leveraging the limited number of values within each factor,

HPSS introduced a heuristic function to guide the search process, assigning higher exploration probability to more promising values, which enhances the effectiveness of mutation and leads to better performance compared to random search. 
We optimize the prompting strategy for two different LLM evaluators (i.e., GPT-4o-mini \cite{openai2023gpt4} and Qwen2.5-14B-Instruct \cite{qwen2}) across four evaluation tasks for natural language generation (NLG). 
Experiment results show that HPSS achieves substantial improvement over the baseline of human-designed evaluation prompts and existing automatic prompt optimization methods. 
Compared to the commonly-used evaluation prompt from MT-Bench \cite{zheng2023judging}, HPSS can achieve an average 29.4\% relative performance improvement with the same generation times. 
Furthermore, compared to other manually designed prompting strategies such as G-Eval \cite{liu-etal-2023-g} and CloserLook \cite{chiang-lee-2023-closer}, HPSS can still achieve significantly better performance with only 5\% of the generation times. 
Our contributions can be summarized as follows:

\begin{itemize}
    \item To the best of our knowledge, we present the first discussion of automatic prompting strategy optimization for LLM evaluators, demonstrating that appropriate prompting strategies can significantly enhance the performance of LLM evaluators.
    \item We propose HPSS, a novel prompting strategy optimization algorithm via iterative search. 
    Furthermore, a heuristic function is introduced to estimate the prospect of each value and enhance the effectiveness of mutation.
    \item We validate the effectiveness of HPSS across four evaluation tasks, yielding consistently better performance compared to both human-designed evaluation prompts and existing automatic prompt optimization methods.
\end{itemize}

\section{Related Work}

\subsection{Prompt Design for LLM Evaluators}
With the emergence of LLMs, utilizing LLMs as evaluators to assess the quality of given texts has gradually become prevalent \cite{wang-etal-2023-chatgpt, chen-etal-2023-exploring-use, ke-etal-2023-decompeval} due to its advantages in flexibility, interpretability, and generalization.
To enhance their performance, 
recent researches attempt to optimize the evaluation prompts for LLM evaluators from multiple aspects
\cite{stureborg2024large, kim-etal-2023-better, jain-etal-2023-multi, murugadoss2024evaluating, he-etal-2024-socreval, pereira2024check}. 
Specifically, G-Eval \cite{liu-etal-2023-g} requires LLMs to generate evaluation steps first and refer to these steps to finish the evaluation task. 
LLMBar \cite{zeng2024llmbar} explores the effects of self-generated reference answers and sample-specific metrics on text evaluation. 
HD-Eval \cite{liu-etal-2024-hd} attempts to align LLM evaluators with humans through hierarchical evaluation criteria decomposition.
However, these works are constrained to optimize an individual factor for evaluation prompts, leading to insufficient optimization. In contrast, our work focuses on adjusting multiple factors and searching for effective prompting strategies to fully stimulate the potential of LLM evaluators.

\subsection{Prompt Optimization}

\begin{table*} [t]
\small
\centering
\resizebox{\linewidth}{!} {
\begin{tabular}{p{0.09\linewidth}p{0.24\linewidth}p{0.44\linewidth}p{0.23\linewidth}}
\toprule
\textbf{Factor} & \textbf{Definition} & \textbf{Common Usage}  & \textbf{Selection Range} \\
\midrule
Scoring \newline Scale & The scoring range for text evaluation. &
Various scoring scales are used in previous work, such as 1-3 \cite{gopalakrishnan2019topical, lin2023llm}, 1-5 \cite{fabbri-etal-2021-summeval, chhun-etal-2022-human}, 1-10 \cite{liu-etal-2024-alignbench} and 1-100 \cite{stureborg2024large}. & 
\begin{minipage}[t]{\linewidth}\vspace{-7pt}\begin{itemize}[noitemsep, topsep=0pt, left=0pt]
    \item \textbf{1-3}
    \item \textbf{1-5}
    \item \textbf{1-10}
    \item \textbf{1-50}
    \item \textbf{1-100}
\end{itemize}\vspace{1pt}\end{minipage} \\
\midrule
In-Context \newline Example & Human evaluation examples & 
Previous works try to construct in-context examples through random or stratified sampling based on human ratings to enhance the performance of LLM evaluators \cite{kim-etal-2023-better, huang2024empirical, jain-etal-2023-multi}. The number of examples usually ranges from 1 to 4. & \begin{minipage}[t]{\linewidth}\vspace{-7pt}\begin{itemize}[noitemsep, topsep=0pt, left=0pt]
    \item \textbf{0 examples}
    \item \textbf{3 examples}
    \item \textbf{5 examples}
    \item \textbf{10 examples}
\end{itemize}\vspace{-7pt}\end{minipage} \\
\midrule
Evaluation \newline Criteria & The general definition of the evaluation aspect and the scoring standards of the quality of a text. & 
The majority of previous works use human-written criteria. Some recent works also prompt LLM evaluators to generate criteria by themselves \cite{kotonya-etal-2023-little} or even use no criteria \cite{murugadoss2024evaluating}. 
&  \begin{minipage}[t]{\linewidth}\vspace{-7pt}\begin{itemize}[itemsep=0.5pt, topsep=0pt, left=0pt]
    \item \textbf{No Criteria}
    \item \textbf{Human-Written Criteria}
    \item \textbf{Self-Generated Criteria}
\end{itemize}\vspace{-7pt}\end{minipage} \\
\midrule
Reference & The reference answer for the task in the sample to evaluate. & 
In addition to using no reference \cite{liu-etal-2023-g}, previous work also utilizes the following types of references: 
(1) \textbf{Human-Written Reference}\footnotemark[1] \cite{zheng2023judging, doddapaneni-etal-2024-finding}, 
(2) \textbf{Self-Generated Reference}: prompting the evaluator to generate a reference independently and then feed it into $ \mathcal{T} $ \cite{zeng2024llmbar}, 
(3) \textbf{Dialectic}: prompting the evaluator to generate a reference along with the final rating \cite{he-etal-2024-socreval}
 & \begin{minipage}[t]{\linewidth}\vspace{-7pt}\begin{itemize}[itemsep=0.5pt, topsep=0pt, left=0pt]
     \item \textbf{No Reference}
     \item \textbf{Self-Generated Reference}
     \item \textbf{Dialectic}
\end{itemize}\vspace{0pt}\end{minipage} 
\\
\midrule
 Chain-of-Thought\newline(CoT) & Request for the LLM evaluator to generate evaluation explanations & Previous work mainly utilizes following CoT formats: (1) \textbf{No CoT}: generate the rating without an explanation \cite{liu-etal-2023-g}, (2) \textbf{Prefix CoT}: provide an explanation first and then give the rating \cite{zheng2023judging, chiang-lee-2023-closer}, and (3) \textbf{Suffix CoT}: provide the rating first, followed by an explanation \cite{chiang-lee-2023-closer} & \begin{minipage}[t]{\linewidth}\vspace{-7pt}\begin{itemize}[itemsep=0.5pt, topsep=0pt, left=0pt]
     \item \textbf{No CoT}
     \item \textbf{Prefix CoT}
     \item \textbf{Suffix CoT}
 \end{itemize}\vspace{-7pt}\end{minipage} \\
 \midrule
 AutoCoT & Self-generated evaluation steps for specific evaluation aspect & G-Eval \cite{liu-etal-2023-g}, OP-I-Prompt \cite{siledar-etal-2024-one} & \begin{minipage}[t]{\linewidth}\vspace{-7pt}\begin{itemize}[noitemsep, topsep=0pt, left=0pt]
     \item \textbf{No AutoCoT}
     \item \textbf{AutoCoT}
 \end{itemize}\vspace{-7pt}\end{minipage} \\
 \midrule
 Metrics & Self-generated sample-specific metrics for good answer & LLMBar \cite{zeng2024llmbar}, Check-Eval \cite{pereira2024check}  & \begin{minipage}[t]{\linewidth}\vspace{-7pt}\begin{itemize}[noitemsep, topsep=0pt, left=0pt]
     \item \textbf{No Metrics}
     \item \textbf{Metrics}
 \end{itemize}\vspace{-7pt}\end{minipage} \\
 \midrule
 Order & The placement order of each component in $ \mathcal{T} $ & For three main components of $ \mathcal{T} $, i.e., Task Description (\textbf{TD}), Evaluation Rule (\textbf{ER}), and Input Content (\textbf{IC}), there are two commonly-used placement orders: \textbf{TD $\rightarrow$ ER $\rightarrow$ IC} \cite{zheng2023judging, liu-etal-2023-g} and \textbf{TD $\rightarrow$ IC $\rightarrow$ ER} \cite{liu-etal-2024-calibrating}.
 & \begin{minipage}[t]{\linewidth}\vspace{-7pt}\begin{itemize}[noitemsep, topsep=0pt, left=0pt]
     \item \textbf{TD $\rightarrow$ ER $\rightarrow$ IC}
     \item \textbf{TD $\rightarrow$ IC $\rightarrow$ ER}
     \item \textbf{ER $\rightarrow$ TD $\rightarrow$ IC}
     \item \textbf{ER $\rightarrow$ IC $\rightarrow$ TD}
     \item \textbf{IC $\rightarrow$ TD $\rightarrow$ ER}
     \item \textbf{IC $\rightarrow$ ER $\rightarrow$ TD}
\end{itemize}\vspace{0pt}\end{minipage} \\
\bottomrule
\end{tabular}}
\vspace{-2mm}
\caption{Factors of evaluation prompts, including their definition, common usage in previous works, and selection range considered in our work. The definitions of each component in the last row are described in Section \ref{factors}. } 
\vspace{-4.5mm}
\label{tab:factors}
\end{table*}

Although LLMs have demonstrated impressive performance across various NLP tasks \cite{zhao2023survey}, researchers point out that their abilities are highly dependent on sophisticated prompt design \cite{pryzant-etal-2023-automatic, leidinger-etal-2023-language, raina-etal-2024-llm}. Considering that manual prompt engineering is time-consuming, many works attempt to optimize prompt automatically. For instance, GPS \cite{xu-etal-2022-gps} and GRIPS \cite{prasad-etal-2023-grips} employ a genetic algorithm, using rule or generative language models to mutate the initial prompt. PromptBreeder \cite{fernando2023promptbreeder} and \textsc{EvoPrompt} \cite{guo2023connecting} utilize LLMs for mutation and crossover operations in evolutionary searches. OPRO \cite{yang2023large}  demonstrates the potential of LLMs to perform iteratively prompt optimization based on search history.
Furthermore, \citet{hsieh-etal-2024-automatic} extends prompt optimization to long prompts containing multiple sentences. They utilized Lin-UCB to guide the selection of sentences and employed in-context learning for LLM-based mutation.
For comparison, our work shifts the focus from word or sentence-level prompt optimization to a higher-level strategy that adjusts multiple factors of evaluation prompts. 
Additionally, we introduce a heuristic function to enhance the effectiveness of mutation based on the characteristics of evaluation prompting strategies.

\footnotetext[1]{Constructing high-quality human-written references requires extensive manual annotation. We do not consider this usage in our study to ensure a fair comparison.}

\section{Methodology}
\begin{figure*}[!t]
\scriptsize
    \centering
    \includegraphics[width=1.0\textwidth]{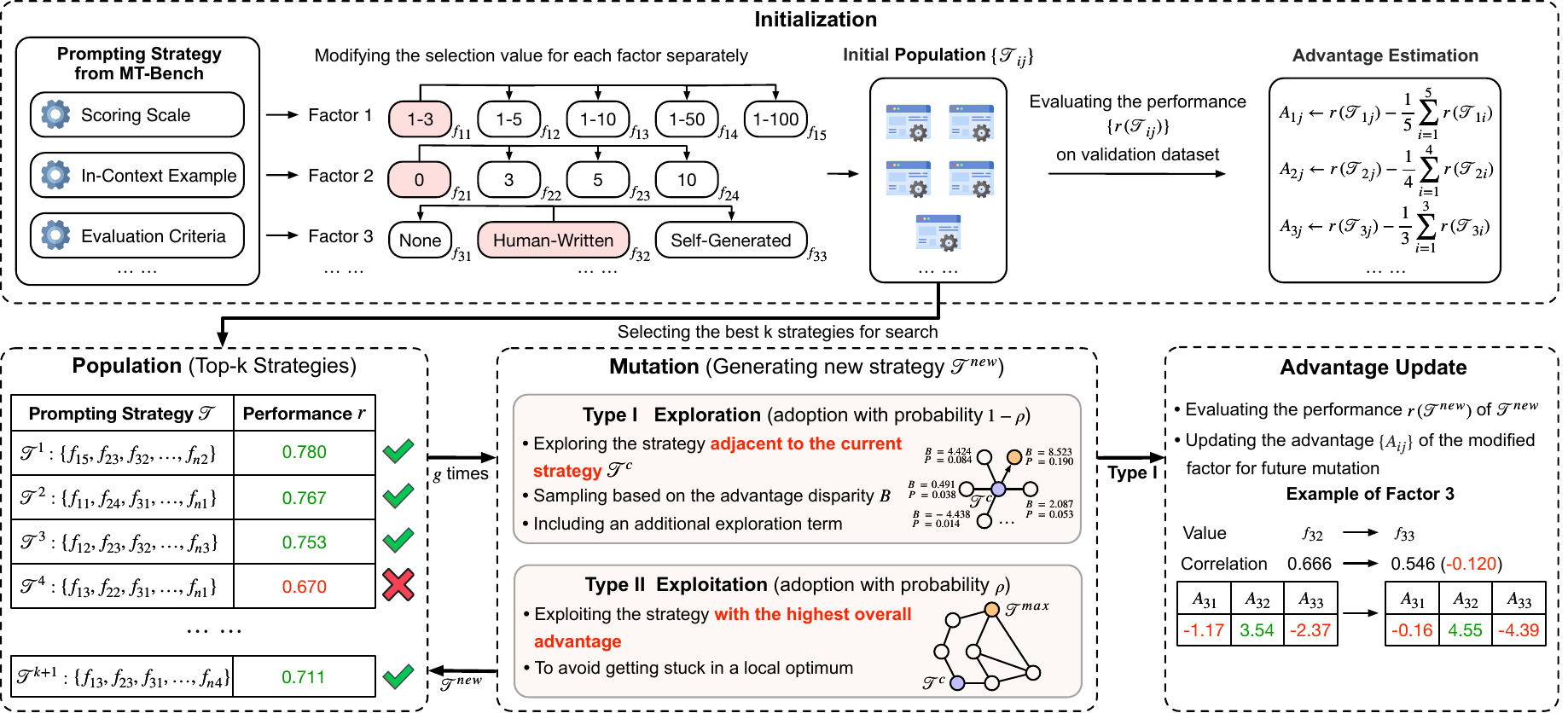}
    \vspace{-4mm}
    \caption{Overview of our HPSS algorithm. By perturbing the value of each factor in the prompting strategy from MT-Bench, HPSS gets the initial strategy population and the advantage estimations of values. Subsequently, new strategies are iteratively searched based on the guidance of the value advantage. The performance $ r $ of new strategies is typically measured by the correlation between the evaluation results of LLM evaluators and human judgments, which is used to update the top-$ k $ strategy population and the advantage estimation.}
    \vspace{-4mm}
    \label{fig:overview}
\end{figure*}

\subsection{Problem Formulation}
NLG evaluation tasks typically require LLM evaluators to generate the corresponding judgment given the input sample $d$ and evaluation aspect $a$. 
The input sample may include the query and model-generated text. 
Meanwhile, the judgment takes various forms, including the rating score of the generated text in pointwise grading and the preference label between two generated texts in pairwise comparison.
To guide LLM evaluators to assess the quality of the input sample $d$ within the aspect $a$, a prompt template $\mathcal{T}$ is needed to provide sufficient clarifications of the task.
In practice, $d$ and $a$ will be fed into $\mathcal{T}$ to constitute an evaluation instruction, and the LLM evaluator will follow this instruction to generate the final judgment.

Since $ \mathcal{T} $ contains various factors, there are multiple prompting strategies for adjusting these factors.
An example of prompting strategy is illustrated in Figure \ref{fig:example}.
Formally, assume that $ \mathcal{T} $ has $ n $ factors $ (F_1, F_2, \ldots, F_n ) $, each factor $ F_{i} $ has $ m_i $ possible values $ (f_{i1}, f_{i2}, \ldots, f_{im_i} ) $. 
Thus, $ \mathcal{T} $ can be determined by $ \boldsymbol{F} = (F_1, F_2, \ldots, F_n) $.
For a dataset $ D=\{d_i\}_{i=1}^{|D|} $ 
and an evaluation aspect $ a $, 
our objective is to search for the prompting strategy $\mathcal{T}^{o}$ to maximize the performance of LLM evaluators:
\begin{equation}
\small
\begin{aligned}
\mathcal{T}^{o} = \arg\max_{\boldsymbol{F}} r_{D,a}(\mathcal{T}_{\boldsymbol{F}})
\end{aligned}
\end{equation}
where 
$ r_{D,a}(\mathcal{T}) $ represents the performance of $ \mathcal{T} $ for the aspect $ a $ on the dataset $ D $,  which is typically measured by the correlation between the evaluation results of LLM evaluators using $ \mathcal{T} $ and human judgments.
Detailed calculation of $ r_{D,a}(\mathcal{T}) $ can be found in Appendix \ref{appendix:correlation}.
To aid the search for the optimal prompting strategy, we will use a validation dataset with human judgments to evaluate the performance of $ \mathcal{T} $.
Without loss of generality, we simplify $ r_{D,a}(\mathcal{T}) $ as $ r(\mathcal{T}) $ to denote the performance of $\mathcal{T}$ for a specific aspect on the validation dataset.

\subsection{Factors of Evaluation Prompt}

\label{factors}

Considering a series of prompting strategies from previous works, we select 8 key factors for the prompting strategy search, as described in Table \ref{tab:factors}. Detailed prompts are provided in Appendix \ref{appendix:prompt_template}.

The first 7 factors belong to 3 main components of the evaluation prompt template $ \mathcal{T} $: 
(1) Task Description (\textbf{TD}), which encompasses the definition of the evaluation task and output format (including \textbf{Scoring Scale} and \textbf{Chain-of-Thought}),
(2) Evaluation Rules (\textbf{ER}), which encompass the standard and steps for evaluation (including \textbf{Evaluation Criteria} and \textbf{AutoCoT}), 
and (3) Input Content (\textbf{IC}), which encompasses the input sample and sample-specific auxiliary information (including \textbf{In-Context Example}, \textbf{Reference}, and \textbf{Metrics}). 
Previous works adopt different orders for these 3 components.
For instance, MT-Bench \cite{zheng2023judging} uses the order \textbf{TD} $\rightarrow$ \textbf{ER} $\rightarrow$ \textbf{IC}, while AutoCalibrate \cite{liu-etal-2024-calibrating} uses \textbf{TD} $\rightarrow$ \textbf{IC} $\rightarrow$ \textbf{ER}. 
We consider the order of these 3 components as the last factor and all 6 possible orders as the selection range.
The overall size of the search space is 12,960. 
We also conduct a preliminary experiment to explore the effect of these factors on Appendix \ref{appendix:preliminary_experiment}.
The results show that these factors significantly influence the performance of LLM evaluators, yet some findings diverge from previous works.

\subsection{Heuristic Prompting Strategy Search}
Inspired by the Genetic Algorithms (GA) \cite{mitchell1980need}, we propose Heuristic Prompting Strategy Search (HPSS) to effectively search prompting strategies to help LLM evaluators achieve better alignment with human judgment. 
The framework is shown in Figure \ref{fig:overview}.
HPSS maintains a population of $k$ top-performing prompting strategies and mutates each strategy in this population to create new strategies. 
Their performance is measured by the human correlation between LLM evaluators and human judgments on the validation dataset.
Different from traditional GA that randomly selects values for mutation with equal probability, 
HPSS leverages the limited number of values within each factor 
and calculates the expected correlation metric advantage of each value over the random selection of the corresponding factor. 
The advantage of value $ f_{ij} $ for factor $ F_i $ denotes as $ A_{ij} $, which can be formally defined as:
\begin{equation}
\small
\begin{aligned}
A_{ij} = \mathop{\mathbb{E}}_{\{F_1, F_2, \ldots, F_n \} \backslash\{F_i\}} [r(\mathcal{T}_{F_1, F_2, \ldots, f_{ij},\ldots, F_n}) \\
-\frac{1}{m_i}\sum_{k=1}^{m_i} r(\mathcal{T}_{F_1, F_2, \ldots, f_{ik},\ldots, F_n}) ]
\end{aligned}
\end{equation}
Then, this advantage score serves as a heuristic function to guide mutation. 
This mechanism assigns higher selection probabilities to more promising values, reducing the cost caused by the blindly random search in GA. 
HPSS consists of two steps:

\definecolor{c1}{RGB}{249,242,234}
\definecolor{c2}{RGB}{228,246,246}
\definecolor{c3}{RGB}{223,243,230}
\definecolor{c4}{RGB}{224,222,241}

\begin{table*} [!t]
\centering
\resizebox{\linewidth}{!} {
\begin{tabular}{cl|ccccc|ccccc}
\toprule
 \multirow{2}{*}{\textbf{Model}} & \multirow{2}{*}{\textbf{Method}}  & \multicolumn{5}{c|}{\textbf{Summeval}}  & \multicolumn{5}{c}{\textbf{Topical-Chat}}  \\
\cmidrule{3-12}
  &  &  Coherence & Consistency & Fluency & Relevance & Avg. & Coherence & Engagingness & Groundedness & Naturalness & Avg.  \\
\midrule
- & BLEU-4  & 0.021 & 0.043 & 0.066 & 0.154 & \cellcolor{c1}0.071 & 0.403 & 0.447 & 0.325 & 0.352 & \cellcolor{c2}0.382  \\
- & \textsc{BertScore}    & 0.228 & 0.194 & 0.225 & 0.370 & \cellcolor{c1}0.254 & 0.390 & 0.438 & 0.351 & 0.335 & \cellcolor{c2}0.379  \\
- & \textsc{UniEval}   & \textbf{0.623} & 0.469 & 0.455 & 0.426 & \cellcolor{c1}0.493 & 0.589 & 0.597 & 0.553 & 0.487 & \cellcolor{c2}0.557  \\
\midrule
\multirow{10}{*}{GPT-4o-mini} & MT-Bench  & 0.481 & 0.482 & 0.319 & 0.445 & \cellcolor{c1}0.432 & 0.546 & 0.635 & 0.530 & 0.536 & \cellcolor{c2}0.562  \\
\cmidrule{2-12}
 & G-Eval$^{\dag}$  & 0.517 & 0.532 & 0.315 & 0.517 & \cellcolor{c1}0.470 & 0.576 & 0.669 & 0.634 & 0.579 & \cellcolor{c2}0.615 \\
 & CloserLook$^{\dag}$  & 0.581 & 0.506 & 0.431 & 0.532 &\cellcolor{c1}0.513 & 0.551 & 0.705 & 0.630 & 0.655 & \cellcolor{c2}0.635 \\
 & CloserLook + ICL  & 0.540 & 0.538 & 0.469 & 0.522 &\cellcolor{c1}0.517 & 0.540 & 0.676 & 0.619 & 0.564 & \cellcolor{c2}0.600 \\
 & CloserLook + ICL$^{\dag}$  & 0.554 & 0.514 & 0.522 & 0.539 & \cellcolor{c1}0.532 & 0.592 & 0.740 & 0.652 & 0.655 & \cellcolor{c2}0.660 \\
\cmidrule{2-12}
 & APE  & 0.560 & 0.516 & 0.401 & 0.514 & \cellcolor{c1}0.498 & 0.542 & 0.668 & 0.705 & 0.562 & \cellcolor{c2}0.619 \\
 & OPRO  & 0.529 & 0.495 & 0.493 & 0.532 & \cellcolor{c1}0.512 & 0.641 & 0.727 & 0.792 & 0.618 & \cellcolor{c2}0.695 \\
 & Greedy  & 0.577 & 0.516 & 0.477 & 0.554 & \cellcolor{c1}0.531 & 0.614 & 0.715 & 0.668 & 0.675 & \cellcolor{c2}0.668 \\
 & Stepwise-Greedy & 0.545 & 0.466 & 0.494 & 0.552 & \cellcolor{c1}0.514 & 0.619 & 0.713 & 0.677 & 0.630 & \cellcolor{c2}0.662 \\
 & HPSS (Ours) & 0.601 & \textbf{0.567} & \textbf{0.525} & \textbf{0.561} &\cellcolor{c1}\textbf{0.564} & \textbf{0.661} & \textbf{0.743} & \textbf{0.866} & \textbf{0.690} & \cellcolor{c2}\textbf{0.740} \\
\midrule
\midrule
\multirow{13}{*}{\begin{tabular}{c}
      \multicolumn{1}{c}{Qwen2.5-14B-} \\
      \multicolumn{1}{c}{Instruct}
    \end{tabular}} & MT-Bench & 0.449 & 0.460 & 0.455 & 0.451 & \cellcolor{c1}0.454 & 0.394 & 0.635 & 0.517 & 0.497 & \cellcolor{c2}0.511  \\
\cmidrule{2-12}
 & G-Eval$^{\dag}$  & 0.569 & 0.544 & 0.467 & 0.515 & \cellcolor{c1}0.524 & 0.490 & 0.673 & 0.505 & 0.535 & \cellcolor{c2}0.550 \\
 & CloserLook$^{\dag}$  & 0.602 & 0.502 & 0.476 & 0.530 & \cellcolor{c1}0.528 &  0.510 & 0.706 & 0.646 & 0.555 & \cellcolor{c2}0.604 \\
 & CloserLook + ICL & 0.478 & 0.587 & 0.355 & 0.480 &\cellcolor{c1}0.475 & 0.436 & 0.675 & 0.642 & 0.580 & \cellcolor{c2}0.583 \\
 & CloserLook + ICL$^{\dag}$ & 0.554 & 0.504 & 0.476 & 0.516 &\cellcolor{c1}0.512 & 0.520 & 0.712 & 0.715 & 0.595 & \cellcolor{c2}0.636 \\
\cmidrule{2-12} 
 & APE  & 0.489\scriptnumber{0.011} & 0.494\scriptnumber{0.019} & 0.451\scriptnumber{0.021} & 0.487\scriptnumber{0.025} &\cellcolor{c1}0.480 & 0.466\scriptnumber{0.025} & 0.670\scriptnumber{0.034} & 0.567\scriptnumber{0.039} & 0.591\scriptnumber{0.037} & \cellcolor{c2}0.574 \\
  & OPRO  & 0.479\scriptnumber{0.041} & 0.564\scriptnumber{0.003} & 0.481\scriptnumber{0.003} & 0.498\scriptnumber{0.010} &\cellcolor{c1}0.506 & 0.502\scriptnumber{0.071} & 0.737\scriptnumber{0.000} & 0.692\scriptnumber{0.000} & 0.583\scriptnumber{0.000} & \cellcolor{c2}0.629 \\
  & Greedy  & 0.528\scriptnumber{0.023} & 0.530\scriptnumber{0.028} & 0.489\scriptnumber{0.007} & 0.527\scriptnumber{0.021} &\cellcolor{c1}0.519 & 0.511\scriptnumber{0.002} & 0.728\scriptnumber{0.030} & 0.776\scriptnumber{0.000} & 0.543\scriptnumber{0.034} & \cellcolor{c2}0.640 \\
  & Stepwise-Greedy  & 0.553\scriptnumber{0.000} & 0.516\scriptnumber{0.000} & 0.491\scriptnumber{0.000} & 0.528\scriptnumber{0.000} &\cellcolor{c1}0.522 & 0.586\scriptnumber{0.000} & 0.721\scriptnumber{0.000} & 0.635\scriptnumber{0.000} & 0.601\scriptnumber{0.000} & \cellcolor{c2}0.636 \\
 & HPSS (Ours)  & 0.559\scriptnumber{0.000} & 0.625\scriptnumber{0.007} & \textbf{0.515}\scriptnumber{0.008} & \textbf{0.542}\scriptnumber{0.006} & \cellcolor{c1}\textbf{0.560} & \textbf{0.598}\scriptnumber{0.030} & \textbf{0.760}\scriptnumber{0.014} & \textbf{0.795}\scriptnumber{0.016} & \textbf{0.635}\scriptnumber{0.017} & \cellcolor{c2}\textbf{0.697} \\
 & - \textit{w/o exploitation}  & 0.543\scriptnumber{0.012} & \textbf{0.626}\scriptnumber{0.008} & 0.511\scriptnumber{0.009} & 0.538\scriptnumber{0.003} & \cellcolor{c1}0.555 & 0.537\scriptnumber{0.038} & 0.741\scriptnumber{0.015} & 0.782\scriptnumber{0.009} & 0.596\scriptnumber{0.021} & \cellcolor{c2}0.664 \\
 & - \textit{w/o exploration term}  & 0.550\scriptnumber{0.012} & 0.623\scriptnumber{0.012} & 0.508\scriptnumber{0.005} & 0.527\scriptnumber{0.008} & \cellcolor{c1}0.552 & 0.555\scriptnumber{0.026} & \textbf{0.760}\scriptnumber{0.013} & 0.789\scriptnumber{0.018} & 0.611\scriptnumber{0.000} & \cellcolor{c2}0.679 \\
 & - \textit{w/o all} & 0.525\scriptnumber{0.010} & 0.617\scriptnumber{0.004} & 0.499\scriptnumber{0.005} & 0.536\scriptnumber{0.002} &\cellcolor{c1}0.544 & 0.516\scriptnumber{0.026} & 0.745\scriptnumber{0.021} & 0.769\scriptnumber{0.005} & 0.568\scriptnumber{0.002} & \cellcolor{c2}0.650 \\
 
\bottomrule
\end{tabular}}
\vspace{-2mm}
\caption{Summary-level Spearman correlations of different aspects on Summeval and dataset-level Spearman correlations on Topical-Chat. \dag \  indicates that the corresponding method employs 20 generations with self-consistency.} 
\label{tab:main_results_1}
\vspace{-4mm}
\end{table*}
\paragraph{Initialization.} 
This step aims to get the initial strategy population and advantage estimation of values.
Starting with a commonly-used reference-free pointwise grading prompting strategy from MT-Bench, we modify the selection value for each factor separately to construct multiple strategies.
After that, we calculate the initial advantage estimation of each value based on its performance and select $k$ top-performing strategies for the subsequent iterative search step.

\paragraph{Iterative Search.}
In each iteration, HPSS mutates each prompting strategy $\mathcal{T}^{c}$ within the population for $g$ times to generate new strategies, 
and then updates the top-$k$ strategy population based on the performance of these new strategies.
To enhance the effectiveness of mutation, HPSS takes overall advantage of all factor values as the performance estimation of one strategy and allocates more search resources to the strategies with higher overall advantages.
Specifically, we utilize two types of mutations: \textit{exploration} and \textit{exploitation}, which have the probability $\rho$ and $ 1 - \rho$ to be chosen for each mutation, respectively.

\noindent (1) \textit{Exploration}: Exploration selects the strategies adjacent to $\mathcal{T}^{c}$ (i.e., the strategies set $T_{adj}$ which can be obtained by modifying a single factor of $\mathcal{T}^{c}$) and
calculates the disparity between the overall advantages of these strategies and $\mathcal{T}^{c}$, employing a temperature-controlled softmax function to determine the exploration probability.
Furthermore, in the preliminary experiment, we notice that inaccurate advantage estimation during \textbf{Initialization} stage may lead to insufficient exploration of well-performing values.
To address this, motivated by UCB Sampling \cite{lai1985asymptotically}, we include an additional exploration term in softmax sampling to incentivize a more comprehensive exploration of all the values. 
Formally, assume the value for $F_i$ within $\mathcal{T}^{c}$ is $f_{ic_i}$, and the strategy $\mathcal{T}$ in $T_{adj}$ modifies the value for $F_i$ from $f_{ic_i}$ to $f_{ij}$, then the exploration probability is computed as follows:                             
\begin{equation}
\small
B_{\mathcal{T}} = \underbrace{A_{ij} - A_{ic_{i}}}_{\text{advantage disparity}} + \underbrace{\lambda\sqrt{\ln(t) / {M_{ij}}}}_{\text{additional exploration term}}
\end{equation}
\begin{equation}
\small
\label{sample}
P(\mathcal{T}) = \frac{\exp(B_{\mathcal{T}} / \tau)}{\sum_{\mathcal{T}' \in T_{adj}} \exp(B_{\mathcal{T}'} / \tau)}
\end{equation}
where $t$ is the search step, and $M_{ij}$ is the appearance count of value $f_{ij}$ during search process.

After exploring a new prompting strategy $\mathcal{T}^{new}$, we evaluate its performance on the validation dataset and calculate the performance gain brought by the modified value. 
After that, a moving average with previous results is applied to update the value's advantage estimation. 
The advantages of the values that belong to the same factor as the modified value will then be normalized to have a zero mean, so as to satisfy the property that the expectation sums to zero.
Formally, the advantage $A$ will be updated as follows, where $N_{ij}$ is the exploration count of the value $f_{ij}$:
\begin{equation}
\label{update_1}
\small
A_{ij} \leftarrow \frac{A_{ij} \cdot N_{ij} + [r(\mathcal{T}^{new}) - (r(\mathcal{T}^{c}) - A_{ic_{i}})]}{N_{ij} + 1}
\end{equation}
\begin{equation}
\small
\label{update_2}
\begin{aligned}
& A_{ik} \leftarrow A_{ik} - \frac{1}{m_i}\sum^{m_i}_{j=1}A_{ij} \\
& \hspace{5.em} k=1,2,\cdots,m_i
\end{aligned}
\end{equation}

\noindent (2) \textit{Exploitation}: 
Since the search scope of \textit{exploration} is constrained to the vicinity of explored prompting strategies, HPSS also employs the mutation type \textit{exploitation} to expand the search scope and avoid getting stuck in a local optimum. 
This method selects the unexplored strategy $\mathcal{T}^{max}$ with the highest overall advantage:
\begin{equation}
\label{exploitation}
\small
\mathcal{T}^{max} = \mathop{\arg\max}\limits_{\boldsymbol{F}}\sum_{i=1}^{n} A_{F_i}
\end{equation}
After reaching the maximum search step, HPSS returns the best-performing strategy on the validation dataset as the final result. 
Detailed implementation of these two steps is presented in Appendix \ref{appendix:algorithm}.

\section{Experiments}
\label{sec:experiments}

\subsection{Experimental Setup}
\paragraph{Tasks and Datasets.}
We evaluate HPSS on four pointwise grading evaluation tasks: Summeval \cite{fabbri-etal-2021-summeval} for text summarization, Topical-Chat \cite{gopalakrishnan2019topical} for dialogue generation, SFHOT / SFRES \cite{wen-etal-2015-semantically} for data-to-text generation, and HANNA \cite{chhun-etal-2022-human} for story generation. 
The Spearman ($\rho$) correlation coefficient between human judgments and LLM evaluations is adopted as the performance metric. 
We also validate HPSS on pairwise comparison benchmarks MT-Bench \cite{zheng2023judging}, AUTO-J \cite{li2024generative}, and LLMBar \cite{zeng2024llmbar}. 
The results are provided in Appendix \ref{appendix:pairwise}.
More details about benchmarks and metrics can be found in Appendix \ref{appendix:benchmarks} and \ref{appendix:correlation}.
Following HD-Eval \cite{liu-etal-2024-hd}, for each dataset, a 50\% proportion is held out for testing, while the rest is applied for validation.

\paragraph{Baselines.}
We compare HPSS with three types of baselines: 
(1) \textbf{Non-LLM Evaluators}: This category includes BLEU-4 \cite{10.3115/1073083.1073135}, \textsc{BertScore} \cite{bert-score}, 
and \textsc{UniEval} \cite{zhong-etal-2022-towards}.
(2) \textbf{Human-Designed LLM Evaluators}: 
The prompting strategy from MT-Bench \cite{zheng2023judging} 
stands as the starting point of searching.
G-Eval \cite{liu-etal-2023-g} integrates AutoCoT to enhance the performance of LLM evaluators. 
\citet{chiang-lee-2023-closer} explores various evaluation schemes. We use its best setting \textit{analyze-rate}, denoted as CloserLook, and attempt to further improve its performance by providing human evaluation examples (CloserLook + ICL). 
(3) \textbf{Automatic Prompt Optimization for LLM Evaluators}: 
We implement 4 strong prompt optimization baselines, including APE \cite{zhou2023large}, 
OPRO \cite{yang2023large}, 
Greedy \cite{prasad-etal-2023-grips, zhou-etal-2023-survival} and Stepwise-Greedy. 
Greedy is a widely used iterative search algorithm, where random perturbations are applied to the current strategy to generate multiple candidates at each iteration, and the best-performing one is retained for the next iteration. 
Stepwise-Greedy sequentially optimizes each factor, choosing the best-performing value in each step.
More details can be found in Appendix \ref{appendix:baselines}.

\definecolor{c1}{RGB}{249,242,234}
\definecolor{c2}{RGB}{228,246,246}
\definecolor{c4}{RGB}{223,243,230}
\definecolor{c3}{RGB}{224,222,241}

\begin{table*} [!t]
\centering
\resizebox{\linewidth}{!} {
\begin{tabular}{cl|ccc|ccc|ccccccc}
\toprule
 \multirow{2}{*}{\textbf{Model}} & \multirow{2}{*}{\textbf{Method}} & \multicolumn{3}{c|}{\textbf{SFHOT}}  & \multicolumn{3}{c|}{\textbf{SFRES}} & \multicolumn{7}{c}{\textbf{HANNA}} \\
\cmidrule{3-15}
  &  & Informativeness & Naturalness & Avg. & Informativeness & Naturalness & Avg. & Relevance & Coherence & Empathy & Surprise & Engagement & Complexity & Avg. \\
\midrule
 - & BLEU-4 & 0.046 & 0.029 & \cellcolor{c1}0.038 & 0.227 & 0.142 & \cellcolor{c2}0.185 & 0.308 & 0.313 & 0.307 & 0.244 & 0.329 & 0.374 & \cellcolor{c3}0.313   \\
 - & \textsc{BertScore}  & 0.103 & 0.076 & \cellcolor{c1}0.090 & 0.213 & 0.150 & \cellcolor{c2}0.182 & 0.298 & 0.383 & 0.403 & 0.331 & 0.387 & 0.436 & \cellcolor{c3}0.373   \\
 - & \textsc{UniEval} & 0.237 & 0.312 & \cellcolor{c1}0.275 & 0.211 & 0.340 & \cellcolor{c2}0.276 & - & - & - & - & - & - & \cellcolor{c3}-   \\
\midrule
\multirow{10}{*}{GPT-4o-mini} & MT-Bench  & 0.307 & 0.385 & \cellcolor{c1}0.346 & 0.253 & 0.417 & \cellcolor{c2}0.335 & 0.499 & 0.457 & 0.475 & 0.365 & 0.490 & 0.500 & \cellcolor{c3}0.464   \\
\cmidrule{2-15} 
 & G-Eval$^{\dag}$  & 0.288 & 0.421 & \cellcolor{c1}0.354 & 0.313 & 0.420 & \cellcolor{c2}0.367 & 0.507 & 0.564 & 0.446 & 0.344 & 0.500 & 0.540 & \cellcolor{c3}0.484 \\
 & CloserLook$^{\dag}$ & 0.325 & 0.439 & \cellcolor{c1}0.382 & 0.312 & 0.395 & \cellcolor{c2}0.353 & \textbf{0.586} & 0.570 & 0.536 & 0.443 & 0.559 & 0.615 & \cellcolor{c3}0.552 \\
& CloserLook + ICL  & 0.327 & 0.445 & \cellcolor{c1}0.386 & 0.295 & 0.379 & \cellcolor{c2}0.337 & 0.528 & 0.518 & 0.493 & 0.384 & 0.552 & 0.575 & \cellcolor{c3}0.508 \\
& CloserLook + ICL$^{\dag}$  & 0.358 & 0.447 & \cellcolor{c1}0.402 & 0.292 & 0.402 & \cellcolor{c2}0.347 & 0.566 & 0.590 & \textbf{0.539} & 0.425 & \textbf{0.585} & \textbf{0.621} & \cellcolor{c3}\textbf{0.555} \\
\cmidrule{2-15} 
 & APE  & 0.289 & 0.427 & \cellcolor{c1}0.358 & 0.272 & 0.373 & \cellcolor{c2}0.323 & 0.524 & 0.494 & 0.507 & 0.321 & 0.427 & 0.499 & \cellcolor{c3}0.462 \\
  & OPRO  & 0.344 & 0.403 & \cellcolor{c1}0.374 & 0.328 & 0.378 & \cellcolor{c2}0.353 & 0.508 & 0.550 & 0.469 & 0.388 & 0.536 & 0.583 & \cellcolor{c3}0.506 \\
  & Greedy  & 0.371 & 0.435 & \cellcolor{c1}0.403 & 0.284 & 0.427 & \cellcolor{c2}0.356 & 0.548 & 0.584 & 0.438 & 0.433 & 0.537 & 0.566 &\cellcolor{c3}0.518 \\
  & Stepwise-Greedy  & 0.368 & 0.452 & \cellcolor{c1}0.410 & 0.292 & 0.382 & \cellcolor{c2}0.337 & 0.566 & 0.517 & 0.522 & 0.424 & 0.537 & 0.521 &\cellcolor{c3}0.515 \\
 & HPSS (Ours)  & \textbf{0.395} & \textbf{0.466} & \cellcolor{c1}\textbf{0.431} & \textbf{0.370} & \textbf{0.439} & \cellcolor{c2}\textbf{0.405} & 0.548 & \textbf{0.594} & 0.535 & \textbf{0.459} & 0.542 & 0.591 & \cellcolor{c3}0.545 \\
\midrule
\midrule
\multirow{13}{*}{\begin{tabular}{c}
      \multicolumn{1}{c}{Qwen2.5-14B-} \\
      \multicolumn{1}{c}{Instruct}
    \end{tabular}} & MT-Bench  & 0.281 & 0.361 & \cellcolor{c1}0.321 & 0.181 & 0.311 & \cellcolor{c2}0.241 & 0.447 & 0.459 & 0.387 & 0.286 & 0.430 & 0.427 & \cellcolor{c3}0.406   \\
\cmidrule{2-15} 
 & G-Eval$^{\dag}$  & 0.262 & 0.418 & \cellcolor{c1}0.340 & 0.308 & \textbf{0.455} & \cellcolor{c2}0.382 & 0.514 & 0.467 & 0.444 & 0.326 & 0.391 & 0.504 & \cellcolor{c3}0.441 \\
 & CloserLook$^{\dag}$ & 0.266 & 0.419 & \cellcolor{c1}0.342 & 0.247 & 0.445 & \cellcolor{c2}0.346 & 0.537 & 0.469 & 0.493 & 0.369 & 0.436 & 0.539 & \cellcolor{c3}0.474 \\
& CloserLook + ICL  & 0.245 & 0.411 & \cellcolor{c1}0.328 & 0.281 & 0.396 & \cellcolor{c2}0.339 & 0.435 & 0.440 & 0.437 & 0.325 & 0.476 & 0.461 & \cellcolor{c3}0.429 \\
& CloserLook + ICL$^{\dag}$  & 0.241 & \textbf{0.419} & \cellcolor{c1}0.330 & 0.336 & 0.406 & \cellcolor{c2}0.371 & 0.560 & 0.532 & \textbf{0.501} & 0.373 & 0.506 & 0.538 & \cellcolor{c3}0.502 \\
\cmidrule{2-15} 
 & APE & 0.310\scriptnumber{0.016} & 0.377\scriptnumber{0.025} & \cellcolor{c1}0.344 & 0.293\scriptnumber{0.033} & 0.426\scriptnumber{0.009} & \cellcolor{c2}0.360 & 0.509\scriptnumber{0.026} & 0.476\scriptnumber{0.004} & 0.456\scriptnumber{0.019} & 0.321\scriptnumber{0.016} & 0.417\scriptnumber{0.018} & 0.450\scriptnumber{0.021} & \cellcolor{c3}0.452 \\
  & OPRO  & 0.301\scriptnumber{0.000} & 0.358\scriptnumber{0.000} & \cellcolor{c1}0.330 & 0.326\scriptnumber{0.025} & 0.388\scriptnumber{0.000} & \cellcolor{c2}0.357 & 0.580\scriptnumber{0.004} & 0.517\scriptnumber{0.015} & 0.436\scriptnumber{0.000} & 0.318\scriptnumber{0.030} & 0.517\scriptnumber{0.009} & 0.550\scriptnumber{0.029} & \cellcolor{c3}0.486 \\
   & Greedy & 0.296\scriptnumber{0.017} & 0.381\scriptnumber{0.016} & \cellcolor{c1}0.339 & 0.294\scriptnumber{0.001} & 0.425\scriptnumber{0.023} & \cellcolor{c2}0.360 & 0.552\scriptnumber{0.011} & 0.493\scriptnumber{0.010} & 0.468\scriptnumber{0.001} & 0.325\scriptnumber{0.034} & 0.495\scriptnumber{0.011} & 0.541\scriptnumber{0.001} & \cellcolor{c3}0.479 \\
    & Stepwise-Greedy & 0.291\scriptnumber{0.000} & 0.345\scriptnumber{0.000} & \cellcolor{c1}0.318 & 0.333\scriptnumber{0.000} & 0.359\scriptnumber{0.000} & \cellcolor{c2}0.346 & 0.506\scriptnumber{0.000} & 0.462\scriptnumber{0.000} & 0.459\scriptnumber{0.000} & 0.383\scriptnumber{0.000} & 0.524\scriptnumber{0.000} & 0.590\scriptnumber{0.000} & \cellcolor{c3}0.487 \\
    & HPSS (Ours)  & \textbf{0.321}\scriptnumber{0.000} & 0.411\scriptnumber{0.000} & \cellcolor{c1}\textbf{0.366} & \textbf{0.351}\scriptnumber{0.005} & 0.444\scriptnumber{0.002} &  \cellcolor{c2}\textbf{0.398} & \textbf{0.586}\scriptnumber{0.004} & \textbf{0.538}\scriptnumber{0.018} & 0.463\scriptnumber{0.005} & \textbf{0.386}\scriptnumber{0.020} & \textbf{0.545}\scriptnumber{0.001} & \textbf{0.594}\scriptnumber{0.007} & \cellcolor{c3}\textbf{0.519} \\
     & - \textit{w/o exploitation}  & 0.295\scriptnumber{0.008} & 0.398\scriptnumber{0.010} & \cellcolor{c1}0.347 & 0.346\scriptnumber{0.006} & 0.425\scriptnumber{0.013} &  \cellcolor{c2}0.386 & 0.544\scriptnumber{0.010} & 0.517\scriptnumber{0.003} & 0.481\scriptnumber{0.017} & 0.370\scriptnumber{0.007} & 0.514\scriptnumber{0.010} & 0.539\scriptnumber{0.004} & \cellcolor{c3}0.494 \\
 & - \textit{w/o exploration term}  & 0.296\scriptnumber{0.006} & 0.403\scriptnumber{0.012} & \cellcolor{c1}0.350 & 0.347\scriptnumber{0.008} & 0.433\scriptnumber{0.011} &  \cellcolor{c2}0.390 & 0.561\scriptnumber{0.027} & 0.516\scriptnumber{0.003} & 0.460\scriptnumber{0.000} & 0.378\scriptnumber{0.012} & 0.521\scriptnumber{0.005} & 0.571\scriptnumber{0.011} & \cellcolor{c3}0.501 \\
  & - \textit{w/o all}  &  0.284\scriptnumber{0.022} & 0.387\scriptnumber{0.014} & \cellcolor{c1}0.336 & 0.348\scriptnumber{0.009} & 0.413\scriptnumber{0.002} & \cellcolor{c2}0.381 & 0.514\scriptnumber{0.012} & 0.524\scriptnumber{0.007} & 0.467\scriptnumber{0.006} & 0.359\scriptnumber{0.002} & 0.483\scriptnumber{0.008}  & 0.521\scriptnumber{0.009} & \cellcolor{c3}0.478 \\
 
\bottomrule
\end{tabular}}
\vspace{-2mm}
\caption{Dataset-level Spearman correlations of different aspects on SFHOT, SFRES, and HANNA. \dag \  indicates that the corresponding method employs 20 generations with self-consistency.} 
\label{tab:main_results_2}
\vspace{-4mm}
\end{table*}
\paragraph{Models and Configurations.}
We choose the representative closed-source model GPT-4o-mini \cite{openai2023gpt4} and open-source model Qwen2.5-14B-Instruct \cite{qwen2} as evaluation models. 
For G-Eval and CloserLook, we use their default settings (with 20 generation times and the decoding temperature as 1.0). 
For other LLM evaluators, greedy search is used for reproducibility. 
We limit the computational budget to 71 evaluations on the validation dataset (including 21 evaluations during initiation in HPSS) for all prompt search methods (21 for Stepwise-Greedy as an exception, as their evaluation count is constant). 
Results of prompt search on Qwen2.5-14B-Instruct are averaged over 3 seeds and the standard deviation is provided.
While for GPT-4o-mini, we report the results of one seed due to the API cost.
More details including the cost of HPSS are in Appendix \ref{appendix:hpss}.

\subsection{Experimental Results}

\paragraph{Human Alignment.}
The main results are presented in Table \ref{tab:main_results_1} and \ref{tab:main_results_2}. 
\textbf{Firstly}, with the same generation times, 
HPSS substantially improves the performance of LLM evaluators compared to the starting point (i.e., the prompting strategy from MT-Bench), resulting in an average 29.4\% relative improvement in Spearman correlation across various tasks and outperforming other Non-LLM and human-designed LLM evaluators by a large margin. 
Even with only 5\% of the generation times, as baseline methods require 20 generations with self-consistency while HPSS only requires a single greedy decoding pass, HPSS remains superior to G-Eval and CloserLook.
Additional experiments in Appendix \ref{appendix:stronger} also demonstrate that the smaller Qwen2.5-14B-Instruct and GPT-4o-mini evaluators with HPSS substantially surpass the larger human-designed Qwen2.5-72B-Instruct and GPT-4o evaluators, respectively.
\textbf{Secondly}, HPSS achieves a significant performance gain over other automatic prompt optimization methods, showing its better adaptability to NLG evaluation tasks.
\textbf{Finally}, HPSS brings consistent performance improvements to different evaluation models, highlighting its cross-model generalizability. 
We also validate that integrating HPSS with inference-time methods such as self-consistency \cite{wang2022self} can further improve the performance of LLM evaluators in Appendix \ref{appendix:sc}. 

\begin{figure}[t]
\scriptsize
    \centering
    \includegraphics[width=1.0\linewidth]{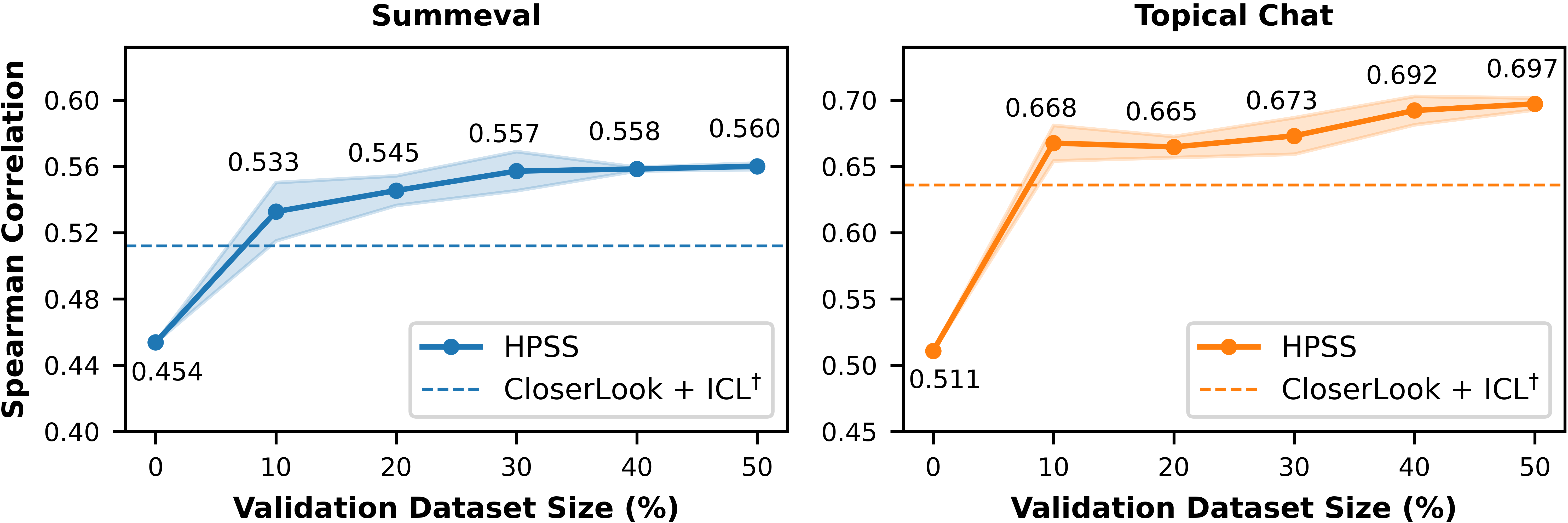}
    \vspace{-4mm}
    \caption{Average performance of Qwen2.5-14B-Instruct evaluator under different validation dataset sizes on Summeval and Topical-Chat.} 
    \vspace{-4mm}
    \label{fig:size}
\end{figure}

\paragraph{Ablation Study.} In Table \ref{tab:main_results_1} and \ref{tab:main_results_2}, we provide an ablation study on key components of HPSS for the Qwen2.5-14B-Instruct evaluator, 
including the mutation type \textit{exploitation}, 
the additional exploration term, 
and the entire heuristic-function-guided mutation mechanism. 
The results validate that all components contribute to the final performance of HPSS. 
Notably, removing the entire mechanism leads to the most significant performance degradation, highlighting its effectiveness.
Moreover, the performance gains from the heuristic search mechanism vary across different datasets, probably due to different search difficulties.

\begin{figure*}[t]
\scriptsize
    \centering
    \includegraphics[width=1.0\textwidth]{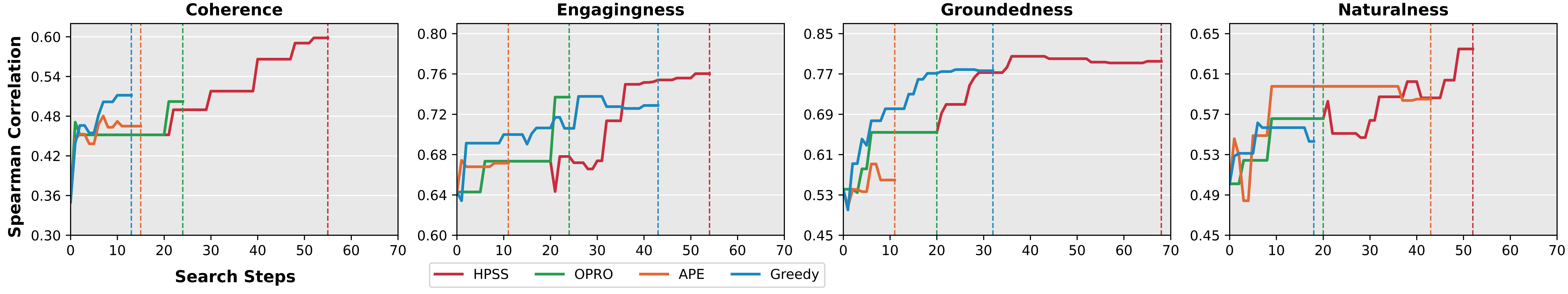}
    \vspace{-4mm}
    \caption{Performance of Qwen2.5-14B-Instruct evaluator under different search steps on Topical-Chat. The first 21 steps for HPSS are in \textbf{Initialization}, which also serves as the initial search history for OPRO, causing the two methods to produce identical results during these steps. Post-convergence results are omitted.} 
    \label{fig:step}
    \vspace{-4mm}
\end{figure*}

\paragraph{The impact of validation dataset size and search budget.} 
Figure \ref{fig:size} shows the performance of HPSS under different validation dataset sizes.
When validating with only 10\% of the human expert annotations, 
the performance of HPSS remains marginally off and superior to human-designed LLM evaluators, 
demonstrating its effectiveness in low-resource scenarios.
Figure \ref{fig:step} shows the performance of Qwen2.5-14B-Instruct evaluator under different search steps on Topical-Chat. 
While HPSS does not exhibit an advantage over baseline methods in the early stages of the search, it avoids getting trapped in local optima prematurely (which often appears in baseline methods) and ultimately converges into better solutions. 
We speculate that this can be attributed to two reasons: 
First, the mutation type \textit{exploitation} expands the search scope of HPSS; 
Second, the advantage estimations for each value become more accurate as the search progresses, enabling an effective search even after good solutions have been obtained. 

\begin{table} [t]
\centering
\resizebox{\linewidth}{!} {
\begin{tabular}{c|c|c|ccc}
\toprule
\multirow{2}{*}{\textbf{Aspects}} & \textbf{Source}  & \textbf{Target} & MT-Bench & CloserLook & \multirow{2}{*}{HPSS}  \\
 & \textbf{Dataset} & \textbf{Dataset} & Prompt & + ICL &  \\
\midrule
\multirow{2}{*}{Coherence} & Summeval  & HANNA   & 0.459 & 0.440 & \textbf{0.516}  \\
 & HANNA  & Summeval   & 0.449 & 0.478 & \textbf{0.505} \\
\midrule
\multirow{2}{*}{Relevance} & Summeval  & HANNA   & 0.447 & 0.434 & \textbf{0.484}          \\
 & HANNA  & Summeval   & 0.451 & 0.480 & \textbf{0.556}        \\
\midrule
\multirow{2}{*}{Engagingness} & Topical-Chat  & HANNA   & 0.430 & 0.476 & \textbf{0.518}         \\
 & HANNA  & Topical-Chat   & 0.635 & 0.675 & \textbf{0.737}        \\
\midrule
\multirow{2}{*}{Naturalness} & Topical-Chat  & SFRES   & 0.311 & 0.396 & \textbf{0.411}         \\
& SFRES  & Topical-Chat   & 0.497 & \textbf{0.580} & 0.566       \\
\bottomrule
\end{tabular}
}
\vspace{-2mm}
\caption{Performance of Qwen2.5-14B-Instruct evaluator when directly applying the prompting strategies found on the source datasets to the target datasets.}
\label{tab:across_datasets}
\vspace{-4mm}
\end{table}
\subsection{Analysis}
\paragraph{The generalizability of prompting strategies across datasets.}
Using Qwen2.5-14B-Instruct as the evaluation model, we directly apply the prompting strategies found by HPSS in Table \ref{tab:main_results_1} and \ref{tab:main_results_2} to other datasets. 
As shown in Table \ref{tab:across_datasets}, these prompting strategies demonstrate strong generalizability across different datasets on the same evaluation aspects, 
as they still achieve significantly better evaluation performances than human-designed prompting strategies in most scenarios. 
We also observe that these prompting strategies show cross-aspect generalization of different datasets. 
However, the improvements in cross-aspect generalization are generally smaller than those in same-aspect generalization, compared to human-designed prompting strategies.
The cross-dataset generalizability of prompting strategies may further reduce the overhead and improve the practicality of HPSS.

\vspace{-0.5mm}
\paragraph{The contribution of each factor.}
\label{app:factor_effect}
\begin{table} [!t]
\small
\centering
\begin{tabular}{l|c|c}
\toprule
\textbf{Factor}  & \textbf{Average $\rho$} & $\Delta$ \\
\midrule
HPSS & \textbf{0.536} & \textbf{0.000} \\
\midrule
- \textit{w/o} Scoring Scale &  0.497 & -0.039 \\
- \textit{w/o} In-Context Example &  0.507 & -0.029 \\
- \textit{w/o} Evaluation Criteria &  0.508 & -0.028 \\
- \textit{w/o} Reference &  0.516 & -0.020 \\
- \textit{w/o} Chain-of-Thought & 0.500 & -0.036 \\
- \textit{w/o} AutoCoT & 0.518 & -0.018 \\
- \textit{w/o} Metrics & 0.527 & -0.009 \\
- \textit{w/o} Order & 0.492 & -0.044 \\
\bottomrule
\end{tabular}
\vspace{-2mm}
\caption{Average performance of HPSS on Summeval, Topical-Chat, SFHOT, and HANNA when each factor is individually removed. Qwen2.5-14B-Instruct is employed as the evaluation model.}
\label{tab:factor_effect}
\vspace{-4mm}
\end{table}
To investigate the contribution of each factor to the overall performance, we conduct an ablation study by independently removing each factor and rerunning HPSS. 
The average performance of Qwen2.5-14B-Instruct evaluator on four datasets is presented in Table \ref{tab:factor_effect}.
The results indicate that each factor contributes to the final performance of HPSS. 
However, the importance of these factors varies. Metrics (i.e., self-generated sample-specific metrics for good answers) appear to be the least important, whereas Order (i.e., the placement order of each component in the evaluation prompt) seems to be the most important. We leave the dedicated search space design to achieve a better trade-off between search overhead and performance as important future work. 

\vspace{-0.5mm}
\paragraph{The advantages of different values.}
We calculate the mean and standard deviations of the advantages for each factor at the end of HPSS across different datasets and evaluation aspects. 
Table \ref{tab:advantages} shows that the advantages exhibit considerable variance across tasks and evaluators, indicating the necessity of finding appropriate prompting strategies tailored to specific tasks and evaluators.
However, there are some common characteristics: 
(1) A moderate scoring scale of 1-10 generally enhances evaluator performance, whereas a too coarse-grained scoring scale of 1-3 is less effective.
(2) Under greedy decoding, directly generating scores and employing human-written evaluation criteria without AutoCoT, Metrics, and Reference generally improves evaluation performance.
(3) Regarding the placement order of different components, positioning the task description at the beginning seems to yield better results, possibly because this arrangement is more logically coherent and helps evaluators concentrate more on sample information and evaluation criteria.
These findings provide insights into future evaluation prompt design.
\definecolor{plus1}{HTML}{cddfee}
\definecolor{plus2}{HTML}{cddfee}
\definecolor{sub1}{HTML}{fff1ea}
\definecolor{sub2}{HTML}{fff1ea}

\begin{table} [t]
\centering
\resizebox{\linewidth}{!} {
\begin{tabular}{cl|c|c}
\toprule
\textbf{Factors} & \textbf{Values} & GPT-4o-mini & Qwen2.5-14B      \\
\midrule
\multirow{5}{*}{
    \begin{tabular}{c}
      \multicolumn{1}{c}{Scoring} \\
      \multicolumn{1}{c}{Scale}
    \end{tabular}
  } & 1-3  & \cellcolor{sub2} -3.545\scriptnumber{1.828}   &\cellcolor{sub2}  -3.500\scriptnumber{2.403}          \\
 & 1-5  & \cellcolor{sub1} -0.310\scriptnumber{1.830}   & \cellcolor{plus1} 0.393\scriptnumber{1.792}          \\
 & 1-10  & \cellcolor{plus1} 1.492\scriptnumber{0.894}   & \cellcolor{plus1} 1.967\scriptnumber{1.652}          \\
 & 1-50  & \cellcolor{plus1} 0.152\scriptnumber{1.456}   & \cellcolor{plus1} 0.462\scriptnumber{1.259}          \\
 & 1-100  & \cellcolor{plus2} 2.211\scriptnumber{0.936}   & \cellcolor{plus1} 0.678\scriptnumber{1.994}          \\
\midrule
\multirow{4}{*}{
    \begin{tabular}{c}
      \multicolumn{1}{c}{In-Context} \\
      \multicolumn{1}{c}{Example}
    \end{tabular}
  } & 0  & \cellcolor{sub2} -2.208\scriptnumber{2.897}   &  \cellcolor{plus1} 0.487\scriptnumber{3.363}    \\
  &  3 & \cellcolor{plus1} 0.715\scriptnumber{2.043}   &  \cellcolor{plus1} 0.860\scriptnumber{3.193}    \\
  &  5 & \cellcolor{sub1} -0.049\scriptnumber{2.826}   &  \cellcolor{plus1} 0.062\scriptnumber{2.913}    \\
  &  10 & \cellcolor{plus1} 1.542\scriptnumber{3.012}   &  \cellcolor{sub1} -1.410\scriptnumber{2.785}    \\
\midrule
\multirow{3}{*}{
    \begin{tabular}{c}
      \multicolumn{1}{c}{Evaluation} \\
      \multicolumn{1}{c}{Criteria}
    \end{tabular}
  } & None  & \cellcolor{sub1} -0.210\scriptnumber{1.849}   &  \cellcolor{sub1} -0.457\scriptnumber{1.936}    \\
   & Human-Written  & \cellcolor{plus1} 0.822\scriptnumber{2.325}   & \cellcolor{plus1} 0.283\scriptnumber{2.158}     \\
  & Self-Generated  & \cellcolor{sub1} -0.612\scriptnumber{1.885}   &  \cellcolor{plus1} 0.174\scriptnumber{1.697}  \\
\midrule
\multirow{3}{*}{Reference} & None  & \cellcolor{plus2} 9.038\scriptnumber{5.772}   & \cellcolor{plus2} 5.513\scriptnumber{4.732}     \\
   & Self-Generated  & \cellcolor{plus2} 3.968\scriptnumber{6.404}   &  \cellcolor{sub1} -1.070\scriptnumber{6.022}     \\
  & Dialectic  & \cellcolor{sub2} -13.00\scriptnumber{11.62} &  \cellcolor{sub2} -4.444\scriptnumber{9.902}  \\
\midrule
\multirow{3}{*}{
    \begin{tabular}{c}
      \multicolumn{1}{c}{Chain-of-} \\
      \multicolumn{1}{c}{Thought}
    \end{tabular}
  } & None  &  \cellcolor{plus1} 0.540\scriptnumber{2.004}  &  \cellcolor{plus1} 0.318\scriptnumber{1.475}    \\
   & Prefix  & \cellcolor{sub1} -0.560\scriptnumber{2.033}   &  \cellcolor{sub1} -0.021\scriptnumber{1.744}     \\
  & Suffix  & \cellcolor{plus1} 0.020\scriptnumber{1.654}   &  \cellcolor{sub1} -0.297\scriptnumber{1.277}  \\
\midrule
\multirow{2}{*}{AutoCoT} & None  & \cellcolor{plus1} 0.308\scriptnumber{0.594}   &  \cellcolor{plus1} 0.346\scriptnumber{1.989}    \\
   & AutoCoT  & \cellcolor{sub1} -0.308\scriptnumber{0.594}   &  \cellcolor{sub1} -0.346\scriptnumber{1.989}     \\
\midrule
\multirow{2}{*}{Metrics} & None  & \cellcolor{plus2} 2.191\scriptnumber{2.575}   &  \cellcolor{plus2} 2.957\scriptnumber{1.892}    \\
   & Metrics  & \cellcolor{sub2} -2.191\scriptnumber{2.575}   &  \cellcolor{sub2} -2.957\scriptnumber{1.892}     \\
\midrule
\multirow{6}{*}{Order} & \textbf{TD $\rightarrow$ ER $\rightarrow$ IC}  & \cellcolor{plus1} 1.128\scriptnumber{1.617}   &  \cellcolor{plus1} 0.201\scriptnumber{2.538}    \\
   & \textbf{TD $\rightarrow$ IC $\rightarrow$ ER}  &  \cellcolor{plus1} 0.408\scriptnumber{2.177}  & \cellcolor{plus1} 1.240\scriptnumber{2.376}      \\
  & \textbf{ER $\rightarrow$ TD $\rightarrow$ IC}  & \cellcolor{plus1} 0.711\scriptnumber{1.970}   & \cellcolor{sub1} -0.393\scriptnumber{2.101}   \\
  & \textbf{ER $\rightarrow$ IC $\rightarrow$ TD}  & \cellcolor{plus1} 0.105\scriptnumber{2.243}   &  \cellcolor{sub1} -0.506\scriptnumber{3.965}  \\
  & \textbf{IC $\rightarrow$ TD $\rightarrow$ ER}  & \cellcolor{sub1} -0.449\scriptnumber{2.254}   &  \cellcolor{plus1} 0.022\scriptnumber{2.335}  \\
  & \textbf{IC $\rightarrow$ ER $\rightarrow$ TD}  & \cellcolor{sub1} -1.902\scriptnumber{1.786}   & \cellcolor{sub1} -0.565\scriptnumber{2.836}  \\
\bottomrule
\end{tabular}
}
\vspace{-2mm}
\caption{Average advantages and standard deviations across different datasets and evaluation aspects for GPT-4o-mini and Qwen2.5-14B evaluators.}
\label{tab:advantages}
\vspace{-4mm}
\end{table}

\begin{figure}[t]
\scriptsize
    \centering
    \includegraphics[width=1.0\linewidth]{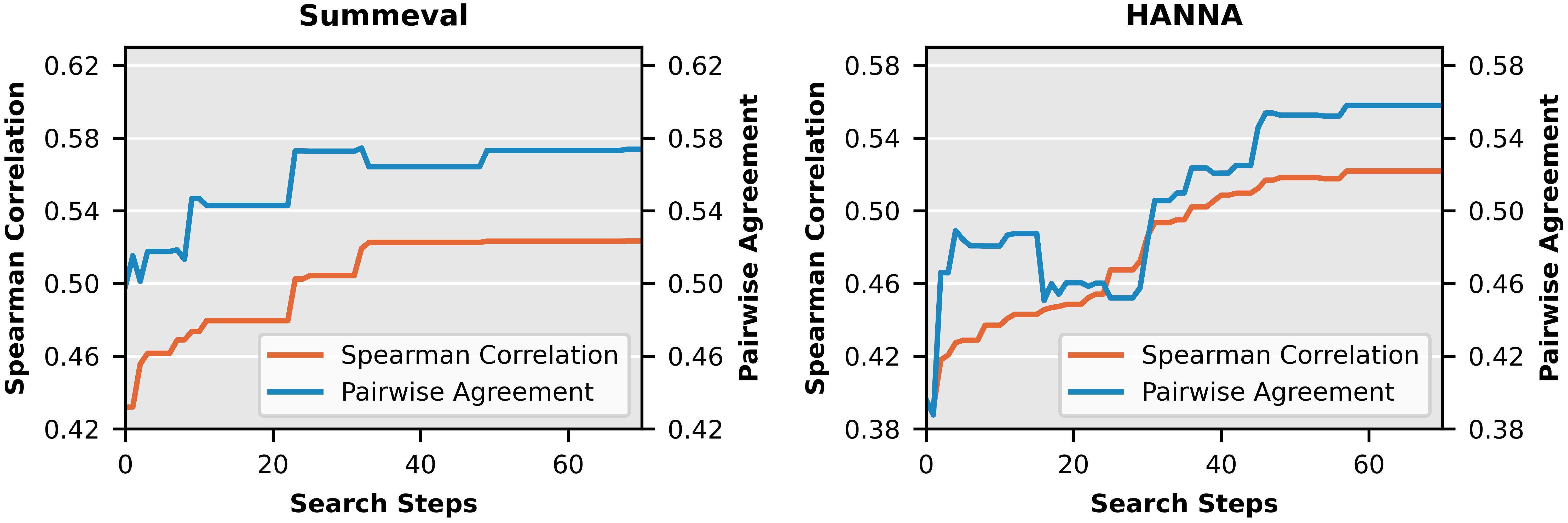}
    \vspace{-4mm}
    \caption{Spearman Correlation and Pairwise Agreement of Qwen2.5-14B-Instruct across different HPSS search steps on the validation datasets of Summeval and HANNA. The first 21 steps are in \textbf{Initialization}, which may cause fluctuations in the metrics.} 
    \vspace{-4mm}
    \label{fig:optimization}
\end{figure}

\vspace{-0.5mm}
\paragraph{Analysis on reward hacking.}
Reward hacking is a widespread problem in the optimization process \cite{skalse2022defining}. 
To analyze this problem in the HPSS framework, we follow AlignBench \cite{liu-etal-2024-alignbench} and introduce a new metric, \textbf{Pairwise Agreement}, to provide an alternative perspective on evaluator performance during our optimization process.
Specifically, for different model-generated texts to the same query (e.g., different generated summaries of the same article for SummEval), we consider all possible pairs and convert their human-judge and LLM-judge scores into pairwise comparison results (i.e., win, tie, or lose), respectively. 
Then, we measure the agreement between LLM evaluators and humans.
On the validation dataset of Summeval and HANNA, the average performance of the Qwen2.5-14B-Instruct evaluator under different search steps of HPSS are shown in Figure \ref{fig:optimization}.
The results demonstrate that the Spearman correlation and pairwise agreement simultaneously improve during the optimization process and verify that the optimization process of HPSS does not introduce reward hacking but genuinely enhances the performance of LLM evaluators. 
We also provide a case study of HPSS in Appendix \ref{app:case}.

\section{Conclusion}
In this work, we integrate 8 key factors for the prompting strategy of LLM evaluators and propose HPSS, a heuristic search method to automatically optimize the prompting strategy for adjusting these factors. 
HPSS leverages the expected performance advantages of each value as the heuristic function to guide the search process. 
Extensive experiments on four NLG evaluation tasks demonstrate the superiority of HPSS, yielding consistent performance gains over both human-designed evaluation prompts and automatic prompt optimization methods. 
Additionally, we validate the generalizability of prompting strategies found by HPSS and analyze the characteristics of these factors.
Our experimental results may provide new insights into future prompt design for LLM evaluators.

\section{Limitations}
The limitations of our work are summarized as follows:

1) HPSS requires iterative search on a validation dataset with human annotations, introducing additional annotation and inference overheads. However, we believe that it is not a severe problem for the following three reasons:
\textbf{Firstly}, the overall cost is affordable in the vast majority of scenarios. Under the default setting, where 50\% of the dataset is selected as the validation dataset, the search cost for a specific evaluation aspect of one dataset is approximately \$8 for GPT-4o-mini, and approximately 40 minutes using 4 H100 GPUs for Qwen2.5-14B-Instruct (See Appendix \ref{appendix:hpss}).
\textbf{Secondly}, HPSS remains effective in low-resource scenarios. As shown in Figure \ref{fig:size}, even when annotated data is scarce 
(reducing the validation dataset size to 10\% of the entire dataset, $ \sim $ 100 samples per dataset)
, HPSS can still significantly outperform manually designed prompt templates.
\textbf{Finally}, HPSS is efficient during inference on the test dataset, requiring only a single greedy decoding pass for each test sample while still achieving better performance than human-designed LLM evaluators that require 20 generations with self-consistency decoding and much larger human-designed LLM evaluators (See Table \ref{tab:main_results_1}, Table \ref{tab:main_results_2}, and Appendix \ref{appendix:stronger}).
Nonetheless, introducing well-designed early stopping strategies
could potentially reduce the search overheads of HPSS, which is considered as important future work.

2) HPSS mainly focuses on the prompting strategy optimization of the input prompts for LLM evaluators. 
Despite the importance of input prompts, other modules such as decoding and interaction strategies may also benefit the performance of LLM evaluators.
Regarding decoding strategies, our experiments demonstrate that incorporating self-consistency decoding can enhance the performance of HPSS (See Appendix \ref{appendix:sc}).
Regarding multiple LLM evaluators' interaction, in our preliminary experiments, we follow the settings of ChatEval \cite{chan2024chateval} to explore the impact of interaction strategies with multiple Qwen2.5-14B-Instruct evaluators, using the prompting strategies found by HPSS. 
We find that different interaction strategies have negligible impacts on the final evaluation performance, which is similar to the self-consistency method that directly averages the evaluation results of multiple individual LLM evaluators. 
Given that the search space increases significantly in multi-agent interaction scenarios and that it is difficult for LLM evaluators with different output formats to interact directly, we leave the optimization of interaction strategies among multiple LLM evaluators as important future work.

\section*{Acknowledgements}
This work was supported by the National Science Foundation for Distinguished Young Scholars (with No. 62125604). 
We would also like to thank Zhipu AI for sponsoring the computational resources in this work.

\bibliography{custom}

\appendix

\clearpage
\startcontents[sections]
\printcontents[sections]{l}{1}{\setcounter{tocdepth}{2}}

\section{Metric Calculation}
\label{appendix:correlation}
\subsection{Pointwise Grading}
Following previous work \cite{liu-etal-2024-hd, liu-etal-2024-calibrating, zhong-etal-2022-towards}, we adopt dataset-level (sample-level for Summeval as an exception) Spearman ($\rho$) correlation coefficient between human judgments and LLM evaluations to measure the performance of LLM evaluators. 
Given a dataset $\mathcal{D}$, evaluation aspect $a$ and evaluation metric $f(\cdot)$, we could calculate the human correlation of this evaluation metric at either dataset or sample level:
\begin{itemize}
\item \textbf{Dataset Level}: For dataset-level human correlation, we evaluate the correlations on all samples in the dataset, as follows:
\begin{equation}
\small
\begin{aligned}
corr_{dataset}(\{s^*_{i,a}\}_{i=1}^{|D|}, \{s_{i,a}\}_{i=1}^{|D|}) = & \\
\rho([{s}^*_{i,a}, ..., {s}^*_{|D|,a}], [s_{i,a}, ..., s_{|D|,a}])
\end{aligned}
\end{equation}
where $\{s^*_{i,a}\}_{i=1}^{|D|}$ and $\{s_{i,a}\}_{i=1}^{|D|}$ denote the 
evaluation results (for free-text evaluations, scores are extracted via rules as final evaluation results) for the aspect $a$ of dataset $\mathcal{D}$ from human annotations and evaluation metric $f(\cdot)$, respectively. 
\item \textbf{Sample Level}: Assume that the dataset $\mathcal{D}$ consists of $J$ queries where each query has target responses from $M$ diverse systems (with a total of $|D| = M \times J$ samples),
and for sample-level human correlation, we first compute correlations on multiple responses to an individual query (e.g., the summaries from 16 summarization systems on one article for Summeval), then average them across all queries:
\begin{equation}
\small
\begin{aligned}
corr_{sample}(\{s^*_{i,a}\}_{i=1}^{|D|}, \{s_{i,a}\}_{i=1}^{|D|}) = & \\
\frac{1}{J} \sum_{i=1}^{J}(\rho([{s}^*_{i1,a}, ..., {s}^*_{iM,a}], [s_{i1,a}, ..., s_{iM,a}]))
\end{aligned}
\end{equation}
where $s^*_{ij,a}$ and $s_{ij,a}$ denote the evaluation results for the $j$-th response to $i$-th query 
for the aspect $a$ of dataset $\mathcal{D}$, from human annotations and evaluation metric $f(\cdot)$, respectively.
\end{itemize}

The evaluation metric $f(\cdot)$ is the LLM evaluator using a specific prompt template in our implementation, and the calculation for the Spearman correlation coefficient $\rho$ between two vectors of length $n$ is as follows:
\begin{equation}
\small
\begin{aligned}
\rho = 1 - \frac{6 \sum d_i^2}{n(n^2 - 1)}
\end{aligned}
\end{equation}
where $d_i$ represents the difference in the rank of the $i$-th element between two vectors, where the ranks are determined by sorting the elements within their respective vectors in ascending order.

\subsection{Pairwise Comparison}
Following previous work \cite{zeng2024llmbar, lambert2024rewardbench}, we directly adopt accuracy to measure the performance of LLM evaluators in pairwise comparison. Given a dataset $\mathcal{D}$, evaluation aspect $a$ and evaluation metric $f(\cdot)$, the accuracy is calculated as follows:
\begin{equation}
\small
\begin{aligned}
acc(\{s^*_{i,a}\}_{i=1}^{|D|}, \{s_{i,a}\}_{i=1}^{|D|}) = 
\frac{1}{|D|}\mathbb{I}(s^*_{i,a}=s_{i,a}) 
\end{aligned}
\end{equation}

\section{List of Evaluation Prompt Templates}
\label{appendix:prompt_template}
This section lists all prompt templates applied throughout this study, including the prompt templates utilized to generate the final rating (Table \ref{tab:evaluation_prompt_summeval}, \ref{tab:evaluation_prompt_topical_chat}, \ref{tab:evaluation_prompt_sfhot}, \ref{tab:evaluation_prompt_hanna}) and the templates used to generate Reference, AutoCoT, and Metrics (Table \ref{tab:generation_prompt_summeval}, \ref{tab:generation_prompt_topical_chat}, \ref{tab:generation_prompt_sfhot}, \ref{tab:generation_prompt_hanna}). 
For these prompt templates, 
we generally refer to the reference-free single answer pointwise grading prompt from MT-Bench \cite{zheng2023judging}. 
However, due to the contents of some components of this template 
being mixed, we make minor adjustments to the order of some sentences to ensure that the search for the factor Order can be conducted.
We use the prompt template of LLMBar \cite{zeng2024llmbar} to generate Metrics.

\begin{figure*}[!t]
  \centering
  \includegraphics[width=0.95\textwidth]{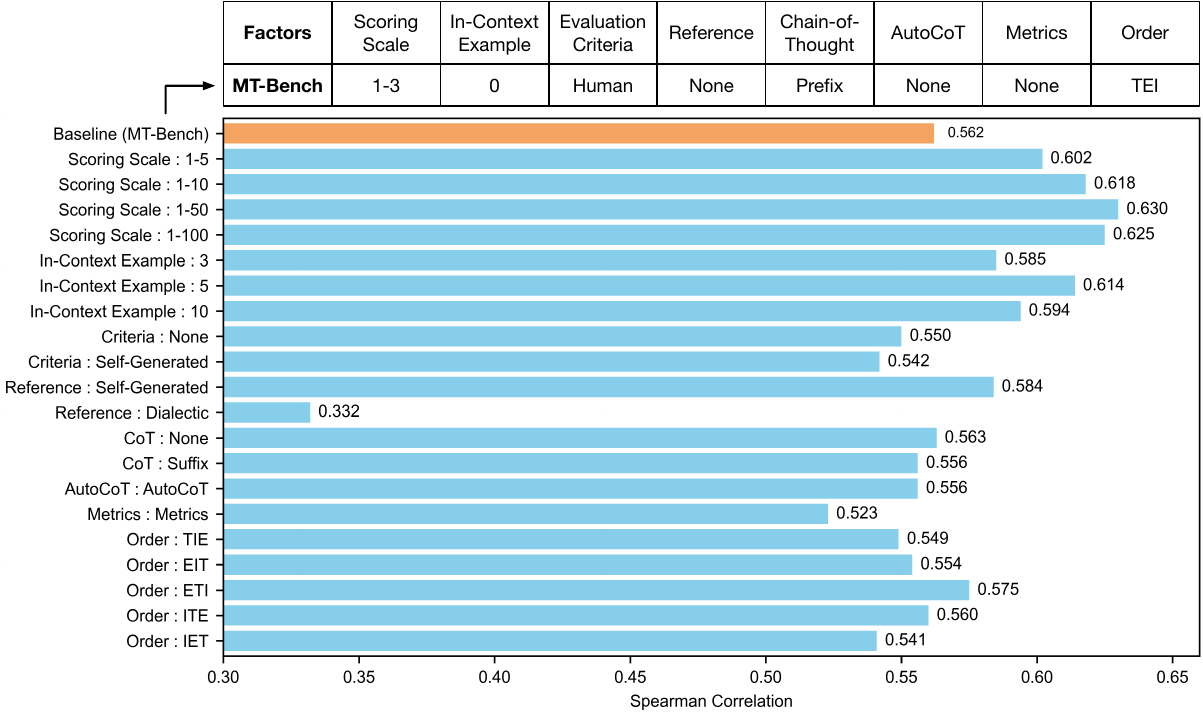}
  \vspace{-2mm}
  \caption{Average dataset-level Spearman human correlation on Topical-Chat for GPT-4o-mini evaluator using different prompting strategies, which are modified based on the baseline strategy from MT-Bench for each factor.}
  \vspace{-4mm}
  \label{fig:init}
\end{figure*}

\section{Preliminary Experiment on Factor Effects}
\label{appendix:preliminary_experiment}
To further explore the effect of each factor on the performance of LLM evaluators, we conducted a preliminary experiment with GPT-4o-mini on a commonly-used dialogue evaluation dataset Topical-Chat \cite{gopalakrishnan2019topical}. 
Using the prompting strategy from MT-Bench \cite{zheng2023judging} with the default scoring scale 1-3 of Topical-Chat as the baseline, we adjust the selection value of each factor separately, employ greedy search decoding  
to generate rating scores, and report the Spearman correlations with human judgments.
The results, as shown in Figure \ref{fig:init}, indicate that these factors significantly influence the performance of LLM evaluators, yet some findings diverge from previous works. 
For instance, \citet{pereira2024check} claims that Metrics can improve the performance of LLM evaluators. 
However, we observed a performance decrease on Topical-Chat. 
\citet{chiang-lee-2023-closer} finds that CoT plays an important role in the evaluation prompt. 
However, when it comes to the GPT-4o-mini evaluator, adding CoT results in negligible differences.
These results highlight the importance of optimizing the prompting strategy for adjusting these factors. 

\section{Details of In-Context Example Selection}
\label{appendix:in_context_example}
We perform stratified sampling based on human ratings within the validation dataset to obtain in-context examples, aiming to ensure an even distribution of examples across different human ratings. 
When evaluating the performance of some prompting strategies on the validation dataset, we remove the corresponding example in the in-context examples set if this example is to be evaluated, aiming to prevent data leakage.

\section{Details of Benchmarks}
\label{appendix:benchmarks}
A brief introduction of the meta-evaluation benchmarks involved is as follows:
\begin{itemize}
\item  \textbf{Summeval} \cite{fabbri-etal-2021-summeval} is a meta-evaluation benchmark for summarization. 
It contains human evaluation annotations for 16 summarization systems on 100 articles from the CNN / DailyMail corpus, resulting in a total of 1600 summary-level annotations. 
Each summary is evaluated on four aspects: \textit{Coherence}, \textit{Consistency}, \textit{Fluency}, and \textit{Relevance}. 
The authors recruit annotators on Amazon Mechanical Turk (AMT) to rate each summary on a scale from 1 to 5. 
Cross-validation by other annotators and experts is conducted to correct errors and enhance annotation quality.
\item \textbf{Topical-Chat} \cite{gopalakrishnan2019topical} is a meta-evaluation benchmark for knowledge-grounded dialogue generation. 
It contains 360 samples, each including dialogue context, relevant knowledge, a response, and human ratings of the response across five aspects: \textit{Coherence}, \textit{Engagingness}, \textit{Groundedness}, \textit{Naturalness}, and \textit{Understandability}, ranging from 1 to 3. 
The annotators recruited from AMT provide the ratings. 
Following HD-Eval \cite{liu-etal-2024-hd}, the first four aspects are used to measure the performance of HPSS.
\item \textbf{SFHOT/ SFRES} \cite{wen-etal-2015-semantically} are meta-evaluation benchmarks for data-to-text generation. 
They contain 875 / 1181 samples respectively, which provide information about restaurants and hotels in San Francisco and aim to let the model generate corresponding utterances. 
The authors recruit annotators from AMT to rate the \textit{Informativeness} and \textit{Naturalness} of the generated utterances for each sample on a scale from 1 to 6.
\item \textbf{HANNA} \cite{chhun-etal-2022-human} serves as a meta-evaluation benchmark for story generation. 
It contains 1,056 stories produced by 10 different automatic story generation systems. 
Each story is rated by 3 annotators recruited from Amazon Mechanical Turk on 6 aspects: \textit{Coherence}, \textit{Relevance}, \textit{Empathy}, \textit{Surprise}, \textit{Engagement}, and \textit{Complexity}. 
Ratings range from 1 to 5. The final score for each aspect is the average of the three annotators' ratings.
\item \textbf{MT-Bench} \cite{zheng2023judging} comprises 3.3k expert-level pairwise human evaluation of responses,  generated by six LLMs on 80 carefully designed questions. 
These questions cover 7 categories: \textit{Writing}, \textit{Roleplay},
\textit{Reasoning Math}, 
\textit{Coding}, \textit{Extraction}, \textit{STEM} and \textit{Humanities}.
We select the first round of dialogues from this dataset and filter out the tied cases, leaving a final evaluation dataset of 1020 instances.

\item \textbf{AUTO-J (Eval-P)} \cite{li2024generative} provides 1,392 pairwise comparison data, each of which contains a query, two LLM-generated responses, and a human-annotated preference label. 
This dataset involves 58 real-world scenarios and the responses are generated from 6 LLM families. 
We filter out the tied cases and leave a final evaluation dataset of 1019 instances.

\item  \textbf{LLMBar} \cite{zeng2024llmbar} is a meta-evaluation benchmark for instruction-following, which consists of two components:
(1) The Natural set, which is gathered from existing human-preference datasets. 
(2) The Adversarial set, where the authors intentionally create misleading outputs that appear plausible but deviate from the instructions to challenge the evaluators.
This dataset contains a total of 419 pairwise comparison instances.

\end{itemize}

\section{Implementation Details}
\subsection{Algorithm Implementation of HPSS}
\label{appendix:algorithm}
We provide the detailed implementation of the two steps of HPSS in Algorithm \ref{algorithm:initiation} and \ref{algorithm:search} respectively.

\subsection{Implementation Details of Baselines}
\label{appendix:baselines}
As for APE, we use the LLM evaluator to resample new prompts. 
The queue size is set to 5. 
In each iteration, two new prompt candidates are resampled based on each prompt in the queue. 
As for OPRO, we use the LLM evaluator to generate new prompting strategies.
We provide the LLM with the selection range for each factor, as well as the 20 previously top-performing strategies and their corresponding correlation metrics, which serve as the search history.  
The LLM is asked to generate a list containing the new selection strategy for each factor. 
The explored prompting strategies in \textbf{Initiation} will serve as the initial search history. 
As for Greedy, we perturb the current prompting strategy 5 times in each iteration by randomly replacing the value of one factor to generate new strategies. 
The strategy that performs best on the validation dataset is retained for the next iteration.
Finally, as for Stepwise-Greedy, we follow the order shown in Table \ref{tab:factors} to optimize each factor sequentially. 
In each step, we select the value for the current factor that performs best on the validation dataset while holding all other factors fixed.
This selected choice is then established as the final optimization result for the current factor.

\subsection{Implementation Details of HPSS}
\label{appendix:hpss}
\begin{figure*}[!h]
\scriptsize
    \centering
    \includegraphics[width=1.0\textwidth]{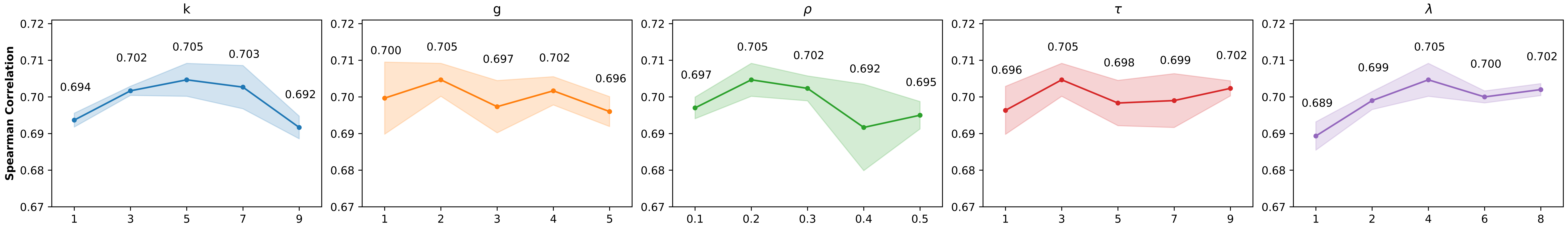}
    \vspace{-4mm}
    \caption{Performance of Qwen2.5-14B-Instruct evaluator under different hyperparameters settings. 
    We provide the average results over 3 seeds on the validation dataset of Topical-Chat.}
    \vspace{-4mm}
    \label{fig:hyper}
\end{figure*}

We determine the hyperparameters for HPSS via grid search on the validation dataset of Topical-Chat using the Qwen2.5-14B-Instruct evaluator. Specifically, the population size $k$ is set to 5.
The mutation time for each template $g = 2$.  The exploitation probability $\rho = 0.2$.
The temperature $\tau$ for the softmax function used to calculate the exploration probability of each template is set to 5, and the weight $\lambda$ for the additional exploration term is set to 4. 
We provide the performance of Qwen2.5-14B-Instruct after modifying each hyperparameter choice on the validation dataset of Topical-Chat in Figure \ref{fig:hyper}.

Under the computational budget described in Section \ref{sec:experiments} (i.e., 71), the search cost of HPSS (i.e., the inference times of the LLM evaluator) for a specific evaluation aspect of one dataset is approximately 70 times the size of the validation dataset. 
When the size of the validation dataset is 50\% of the entire dataset, for GPT-4o-mini, the cost of HPSS for a specific evaluation aspect of one dataset is approximately \$8. 
In total, the overall cost of HPSS across all datasets is approximately \$140. 
For Qwen2.5-14B-Instruct, the runtime of HPSS for a specific evaluation aspect of one dataset is approximately 40 minutes using 4 H100 GPUs and the vllm \cite{10.1145/3600006.3613165} inference framework, while the overall runtime of HPSS across all datasets is approximately 12 hours. 
The overall costs for other automatic prompt optimization methods for LLM evaluators are the same as HPSS, which is affordable in the vast majority of scenarios.
We also validate that even with 1/5 of the above search cost (reducing the validation dataset size to 10\% of the entire dataset), HPSS can still significantly outperform human-designed LLM evaluators in Figure \ref{fig:size}.

\section{Experiments on Pairwise Comparison}
\label{appendix:pairwise}
\subsection{Experimental Setup}
Apart from pointwise grading tasks, we also validate our method on three pairwise comparison benchmarks, i.e., MT-Bench \cite{zheng2023judging}, AUTO-J \cite{li2024generative}, and LLMBar \cite{zeng2024llmbar}, which primarily focus on instruction-following tasks.
Qwen2.5-14B-Instruct is employed as the evaluation model, with all hyperparameters remaining the same as pointwise grading experiments. 
Regarding the search space, we remove the factor Scoring Scale, and the value \textit{Self-Generated Criteria} of the factor Evaluation Criteria, which only exist in the pointwise grading setting. 
We compare our method with the prompting strategies from MT-Bench, the best human-designed prompting strategies \textit{Metrics+Reference$^*$} found by LLMBar, and all automatic prompt optimization methods examined in pointwise grading experiments.
Results of prompt search are averaged over 3 random seeds and the standard deviation is provided.
\subsection{Main Results}
The main results are provided in Table \ref{tab:pairwise_results}.
HPSS substantially improves the performance of LLM evaluators compared to human-designed prompting strategies and achieves the best average performance across all automatic prompt optimization methods, which validates its effectiveness in pairwise comparison prompt optimization.
On the AUTO-J (Eval-P), multiple prompt optimization methods simultaneously achieve the best results.
Upon data examination, we find that 
the best prompting strategy is close to the starting point, with only one factor having a different value, resulting in lower search difficulty. 
Overall, we observe that human-designed prompting strategies for pairwise comparison tasks have already been well-optimized, and the improvements brought by automatic prompt optimization are relatively modest compared to those in pointwise grading tasks.

\begin{table} [t]
\centering
\resizebox{\linewidth}{!} {
\begin{tabular}{l|c|c|c|c}
\toprule
\multirow{2}{*}{\textbf{Method}}  & \textbf{AUTO-J}   & \multirow{2}{*}{\textbf{LLMBar}}  & \multirow{2}{*}{\textbf{MT-Bench}} & \multirow{2}{*}{\textbf{Avg.}} \\
 & \textbf{(Eval-P)} & & & \\
\midrule
 MT-Bench  & 0.792 & 0.619 & 0.765 &  0.725 \\
 Metrics+Reference*  & 0.800 & 0.724 & 0.778 & 0.767\\
\cmidrule{1-5}
APE & 0.799\scriptnumber{0.005} & 0.670\scriptnumber{0.032} & 0.774\scriptnumber{0.008} & 0.748 \\
OPRO & \textbf{0.847}\scriptnumber{0.000} & 0.695\scriptnumber{0.000} & \textbf{0.791}\scriptnumber{0.005} & 0.778 \\
Greedy & 0.820\scriptnumber{0.019} &  0.774\scriptnumber{0.005} & 0.775\scriptnumber{0.013} & 0.790 \\
Stepwise-Greedy  & \textbf{0.847}\scriptnumber{0.000} & 0.743\scriptnumber{0.000} & 0.784\scriptnumber{0.000} & 0.791 \\
HPSS (Ours) & \textbf{0.847}\scriptnumber{0.000} & \textbf{0.778}\scriptnumber{0.005} & 0.789\scriptnumber{0.015} & \textbf{0.805} \\
\bottomrule
\end{tabular}
}
\vspace{-2mm}
\caption{Accuracy of different prompting methods based on Qwen2.5-14B-Instruct on pairwise comparison datasets including AUTO-J (Eval-P), LLMBar, and MT-Bench.}
\label{tab:pairwise_results}
\vspace{-4mm}
\end{table}
\begin{table*} [!t]
\centering
\small
\begin{tabular}{c|l|c|c|c|c|c}
\toprule
\textbf{Model} & \textbf{Method}  & \textbf{Summeval} & \textbf{Topical-Chat} & \textbf{SFHOT} & \textbf{HANNA} & \textbf{Average} \\
\midrule
\multirow{2}{*}{Qwen2.5-72B-Instruct} & MT-Bench  & 0.502  &  0.634  & 0.362  &  0.460 & 0.490  \\
& CloseLook + ICL  &  0.481  & 0.592 & 0.316 & 0.470  & 0.465  \\
\midrule
Qwen2.5-14B-Instruct & HPSS & \textbf{0.560}  &  \textbf{0.697}  &  \textbf{0.366} & \textbf{0.519} & \textbf{0.536} \\
\midrule
\midrule
\multirow{2}{*}{GPT-4o} & MT-Bench  & 0.479  &  0.642  & 0.295  &  0.420 & 0.459 \\
& CloseLook + ICL  &  0.519  & 0.639 & 0.300 & 0.487 &  0.486  \\
\midrule
GPT-4o-mini & HPSS & \textbf{0.564}  &  \textbf{0.740}  &  \textbf{0.431} & \textbf{0.545}  & \textbf{0.570} \\
\bottomrule
\end{tabular}
\vspace{-2mm}
\caption{Average performance comparison of HPSS and human-designed LLM evaluators on Summeval, Topical-Chat, SFHOT, and HANNA. \dag \  indicates that the corresponding method employs 20 generations with self-consistency.}
\label{tab:stronger_model}
\end{table*}
\begin{table*} [!t]
\small
\centering
\begin{tabular}{l|c|c|c|c|c}
\toprule
\textbf{Method}  & \textbf{Summeval} & \textbf{Topical-Chat} &\textbf{SFHOT} & \textbf{HANNA} & \textbf{Average}  \\
\midrule
CloseLook + ICL$^{\dag}$ & 0.532  &  0.660  &  0.402 & 0.555 &  0.537     \\
HPSS & 0.564   &  0.740 & 0.431 & 0.545  &  0.570   \\
HPSS$^{\dag}$ & \textbf{0.569}    & \textbf{0.763} &   \textbf{0.453} & \textbf{0.578}  &   \textbf{0.591}   \\
\bottomrule
\end{tabular}
\vspace{-2mm}
\caption{Average performance of HPSS under different generation times on Summeval, Topical-Chat, SFHOT, and HANNA. \dag \  indicates that the corresponding method employs 20 generations with self-consistency. GPT-4o-mini is employed as the evaluation model.}
\vspace{-2mm}
\label{tab:sc}
\end{table*}
\section{Comparison with Stronger LLM Evaluator}
\label{appendix:stronger}
As shown in Table \ref{tab:stronger_model}, we compare the performance of smaller evaluators (i.e., Qwen2.5-14B-Instruct, GPT-4o-mini) using the prompting strategies obtained by HPSS with that of stronger evaluators (i.e., Qwen2.5-72B-Instruct, GPT-4o) using human-designed prompting strategies. 
Notably, the smaller evaluators with HPSS achieve
significantly better evaluation performance than the human-designed stronger evaluators across different datasets.
These results demonstrate the efficiency of HPSS.

\section{Incorporating HPSS with Inference-Time Methods}
\label{appendix:sc}
As illustrated in Table \ref{tab:main_results_2}, for GPT-4o-mini evaluator, 
the performance of CloserLook + ICL surpasses HPSS on HANNA. 
Considering the different inference overheads of these two methods, 
we attempt to incorporate self-consistency \cite{wang2022self} decoding strategy into HPSS to ensure a fair comparison. 
Specifically, we conduct 20 generations and compute the average evaluation score,
which is consistent with the setting of CloserLook + ICL. 
As shown in Table \ref{tab:sc}, self-consistency further enhances the performance of HPSS, consistently outperforming CloserLook + ICL with the same generation times. 
This result indicates that integrating HPSS with inference-time methods can further improve the performance of LLM evaluators.

\begin{algorithm*}[h]
\caption{Initialization}
\label{algorithm:initiation}
\begin{algorithmic}[1]
\small
\Require $ n $ factors $ F_1, F_2, \ldots, F_n $, baseline prompting strategy $ \mathcal{T}^{base} = \mathcal{T}_{f_{1b_{1}}, f_{2b_{2}}, \ldots, {f_{nb_{n}}}} $, performance metrics $r$
\State $T \gets \{\mathcal{T}^{base}\} $
\State $s_{base} \gets c(\mathcal{T}^{base}) $
\For{$i = 1, \cdots, n$} 
\For{$j = 1, \cdots, m_i \ \text{and} \ j \neq b_i$}
\State $T \gets T \cup \{\mathcal{T}_{f_{1b_{1}}, f_{2b_{2}}, \ldots, f_{ij}, \ldots, {f_{nb_{n}}}}\} $
\State $s_{ij} \gets r(\mathcal{T}_{f_{1b_{1}}, f_{2b_{2}}, \ldots, f_{ij}, \ldots, {f_{nb_{n}}}}) $
\EndFor
\For{$j = 1, \cdots, m_i$}
\State $A_{ij} \gets s_{ij} - \frac{1}{m_i}\sum^{m_i}_{k=1}s_{ik}$
\EndFor
\EndFor
\State \Return the best $k$ prompt templates based on $s$,
as $best$
\end{algorithmic}
\end{algorithm*}

\begin{algorithm*}[h]
\caption{Iterative Search}
\label{algorithm:search}
\begin{algorithmic}[1]
\small
\Require Explored prompting strategies set $T$, initial prompting strategies population $best$, performance metrics $r$, beam size $k$, budget $m$, hyper-parameters $\lambda, g, \tau, \rho$
\State $cost \gets 0 $
\While{$cost < m $}
\State $new\_best \gets best$
\For {each $\mathcal{T}^{c} $ in $best$}
\For {$l = 1, \cdots, g$}

\State Sample new candidate prompting strategy $\mathcal{T}^{new}$ in $T_{adj}$ based on Equation \ref{sample} \Comment{Exploration}
\If {$\mathcal{T}^{new} \in T$} 
\State \textbf{continue} \Comment{Skip explored strategies}
\EndIf
\State Sample $\alpha \sim \text{Bernoulli}(\rho)$
\If {$\alpha$}
\State Select $\mathcal{T}^{max}$ with the highest overall advantage and $\mathcal{T}^{max} \notin T$ based on Equation \ref{exploitation} \Comment{Exploitation}
\State $\mathcal{T}^{new} \gets \mathcal{T}^{max} $
\EndIf
\State $T \gets T \cup \{\mathcal{T}^{new}\} $
\State $s^{new} \gets r(\mathcal{T}^{new})$
\State $new\_best \gets new\_best \cup \{\mathcal{T}_{new}\}$
\State $cost \gets cost + 1$
\If {not $\alpha$}
\State Update advantage $A$ based on Equation \ref{update_1} and \ref{update_2} \Comment{Advantage update}
\EndIf
\EndFor
\EndFor
\State Select the best $k$ prompting strategies in $new\_best$, as $best$
\EndWhile
\State \Return $ best[0] $
\end{algorithmic}
\end{algorithm*}

\section{The Selection of Default Prompting Template}
\label{app:prompt_template}
\begin{table*} [!t]
\small
\centering
\begin{tabular}{l|c|c|c|c|c}
\toprule
\textbf{Prompt Template}  & \textbf{Summeval} & \textbf{Topical-Chat} &\textbf{SFHOT} & \textbf{HANNA} & \textbf{Average}  \\
\midrule
MT-Bench \cite{zheng2023judging} & 0.432  &  0.562  &  \textbf{0.346} & 0.464 & \textbf{0.451} \\
G-Eval \cite{liu-etal-2023-g} & \textbf{0.436}   &  \textbf{0.579} & 0.315 & 0.466 & 0.449 \\
Themis \cite{hu-etal-2024-themis} & 0.396    & 0.531 &   0.322 & \textbf{0.474} & 0.431    \\
\bottomrule
\end{tabular}
\vspace{-2mm}
\caption{Average performance of GPT-4o-mini evaluator under different prompting templates on Summeval, Topical-Chat, SFHOT, and HANNA.}
\vspace{-4mm}
\label{tab:prompt_template}
\end{table*}
We choose the prompt template of MT-Bench as the starting point because it has been widely used in previous works about LLM-as-a-Judge \cite{doddapaneni-etal-2024-finding, zhang2025reviseval, song-etal-2025-many}. 
To verify the effectiveness of this prompt template, we compare the performance of the GPT-4o-mini evaluator under different prompt templates in Table \ref{tab:prompt_template}. We find that the MT-Bench prompt template achieves comparable performance to the best-performed prompt template across various datasets, which validates the rationale for choosing it as the starting point.

\definecolor{customcolor1}{RGB}{255, 231, 218}
\definecolor{customcolor2}{RGB}{218, 233, 248}
\definecolor{customcolor3}{RGB}{255, 245, 213}

\begin{table*} [t]
\scriptsize
\centering
\setlength{\tabcolsep}{1.6mm}{
\begin{tabular}{l l >{\raggedright\arraybackslash}m{0.7\textwidth}}
\toprule
\textbf{Template} & \textbf{Value} & \textbf{Prompt} \\
\midrule
\multirow{5}{*}[-7em]{Backbone} &  \multirow{5}{*}[-7em]{-} & 
\begin{tabular}[t]{@{}>{\cellcolor{customcolor1}}p{0.7\textwidth}}
        \#\# Instruction \newline
        Please act as an impartial judge and evaluate the quality of the summary of the news article displayed below on its \textcolor{blue}{[Aspect]}. \textcolor{red}{\{reference\_1\_template\}} \textcolor{red}{\{reference\_dialectic\_template\}} \textcolor{red}{\{chain\_of\_thought\_template\}} \\
    \end{tabular} \\
& & \  \\
& & 
\begin{tabular}[t]{@{}>{\cellcolor{customcolor2}}p{0.7\textwidth}}
Here are some rules of the evaluation: \newline
1. Your evaluation should consider the \textcolor{blue}{[Aspect]} of the summary. \textcolor{blue}{[Criteria]} \newline
2. Be as objective as possible. \newline\newline
\textcolor{red}{\{autocot\_template\}} \\
\end{tabular} \\
& & \ \\
& &
\begin{tabular}[t]{@{}>{\cellcolor{customcolor3}}p{0.7\textwidth}}
\textcolor{red}{\{in\_context\_example\_template\}} \newline
\#\# Article \newline
\textcolor{blue}{[Article]} \newline\newline
\textcolor{red}{\{metrics\_template\}} \newline\newline
\textcolor{red}{\{reference\_2\_template\}} \newline\newline
\#\# The Start of the Summary \newline
\textcolor{blue}{[Summary]} \newline
\#\# The End of the Summary 
 \\
 \end{tabular} \\
\midrule
\multirow{1}{*}{Reference 1} & - & You will be given the news article, the summary, and a high-quality reference summary. \\
\midrule
\multirow{1}{*}{Reference 2} & - & \#\# The Start of Reference Summary \newline
\textcolor{blue}{[Reference]} \newline
\#\# The End of Reference Summary \\

\midrule
\multirow{1}{*}{Reference Dialectic} & - & Please generate your own summary for the news article first and take into account your own summary to evaluate the quality of the given summary. \\
\midrule
\multirow{3}{*}[-1.85em]{Chain-of-Thought} & No CoT & You must directly output your rating of the summary on a scale of 1 to \textcolor{red}{\{max\}} without any explanation by strictly following this format: "[[rating]]", for example: "Rating: [[\textcolor{red}{\{max\}}]]". \\
\cmidrule{2-3}
 & Prefix CoT & Begin your evaluation by providing a short explanation. After providing your explanation, you must rate the summary on a scale of 1 to \textcolor{red}{\{max\}} by strictly following this format: "[[rating]]", for example: "Rating: [[\textcolor{red}{\{max\}}]]". \\
 \cmidrule{2-3}
 & Suffix CoT & You must rate the summary on a scale of 1 to \textcolor{red}{\{max\}} first by strictly following this format: "[[rating]]", for example: "Rating: [[\textcolor{red}{\{max\}}]]". And then provide your explanation. \\
\midrule
\multirow{5}{*}[-1.5em]{Scoring Scale} & 3 &  \textcolor{red}{\{max\}} $ = 3 $ \\
\cmidrule{2-3}
 & 5 & \textcolor{red}{\{max\}} $ = 5 $ \\
\cmidrule{2-3}
 & 10 & \textcolor{red}{\{max\}} $ = 10 $ \\
\cmidrule{2-3}
 & 50 & \textcolor{red}{\{max\}} $ = 50 $ \\
\cmidrule{2-3}
 & 100 & \textcolor{red}{\{max\}} $ = 100 $ \\
\midrule
\multirow{1}{*}{AutoCoT} & - & Evaluation Steps: \newline
\textcolor{blue}{[Autocot]} \\
\midrule
\multirow{1}{*}{In-Context Example} & - & Here are some examples and their corresponding ratings: \newline
\textcolor{red}{\{example\_template\_1\}} \newline
\textcolor{red}{\{example\_template\_2\}} \newline
\textcolor{red}{…} \newline
\textcolor{red}{\{example\_template\_n\}}\newline\newline
Following these examples, evaluate the quality of the summary of the news article displayed below on its \textcolor{blue}{[Aspect]}: \\
\midrule
Example & - & \#\# Example \textcolor{blue}{[Number]}:\newline \#\# Article\newline\textcolor{blue}{[Article]}\newline\newline \#\# The Start of the Summary \newline\textcolor{blue}{[Summary]}\newline \#\# The End of the Summary\newline\newline \#\# Rating \newline \textcolor{blue}{[Human Rating]}\\
\midrule
\multirow{1}{*}{Metrics} & - & \#\# Questions about Summary \newline
Here are some questions about the summary. You can do the evaluation based on thinking about all the questions. \newline
\textcolor{blue}{[Metrics]} \\
\bottomrule
\end{tabular}
}
\vspace{-2mm}
\caption{Detailed evaluation prompt templates for Summeval. 
The backbone serves as the final input prompt template for LLM evaluators. 
The three components marked in different colors represent Task Description (\textbf{TD}), Evaluation Rule (\textbf{ER}), and Input Content (\textbf{IC}), respectively. 
The content within \textcolor{red}{\{\}} represents the prompt template for each factor, corresponding to the following rows in this table. 
Different content may be chosen for each template when corresponding factor values vary. 
The content within \textcolor{blue}{[]} is sample-specific input information. 
- in \textbf{Value} means that when the factor is chosen as "None", this template will be replaced with an empty string ("").
Otherwise, the content of this template will be added to the backbone. 
Specifically, the templates Reference 1 and Reference 2 will be replaced with an empty string ("") unless the factor Reference is chosen as \textbf{Self-Generated Reference}.
The template Reference Dialectic will be replaced with an empty string ("") unless the factor Reference is chosen as \textbf{Dialectic}.}
\label{tab:evaluation_prompt_summeval}
\end{table*}
\begin{table*} [t]
\centering
\scriptsize
\resizebox{\linewidth}{!} {
\setlength{\tabcolsep}{1.6mm}{
\begin{tabular}{l >{\raggedright\arraybackslash}m{0.85\textwidth}}
\toprule
\textbf{Template} & \textbf{Prompt} \\
\midrule
\multirow{1}{*}{Reference Generation} & 
Please summarize the following text: \textcolor{blue}{[Article]} \newline
Summary:  \\
\midrule
\multirow{1}{*}{AutoCoT Generation} & 
\#\# Instruction \newline
Please act as an impartial judge and evaluate the quality of the summary of the news article on its \textcolor{blue}{[Aspect]} and rate the summary on a scale of 1 to \textcolor{red}{\{max\}}.\newline\newline 
Here are some rules of the evaluation: \newline
1. Your evaluation should consider the \textcolor{blue}{[Aspect]} of the summary. \textcolor{blue}{[Criteria]} \newline
2. Be as objective as possible. \newline\newline
Please generate the evaluation steps for this task without other explanation.\newline
Evaluation Steps: \\
\midrule
\multirow{1}{*}{Metrics Generation} & 
\#\# Instruction \newline
Please act as an impartial judge and evaluate the quality of the summary of the news article displayed below on its \textcolor{blue}{[Aspect]}. Please propose at most three concise questions about whether a potential summary is a good summary for a given news article on its \textcolor{blue}{[Aspect]}. Another assistant will evaluate the aspect of the summary by answering all the questions. \newline

Here are some rules of the evaluation: \newline
(1) Your evaluation should consider the \textcolor{blue}{[Aspect]} of the summary. \textcolor{blue}{[Criteria]} \newline
(2) Outputs should NOT contain more/less than what the instruction asks for, as such outputs do NOT precisely execute the instruction. \newline

\#\# Article: \newline
\textcolor{blue}{[Article]} \newline

\#\# Requirements for Your Output: \newline
(1) The questions should **specifically** target the given news article instead of some general standards so that the questions may revolve around key points of the news article. \newline
(2) You should directly give the questions without any other words. \newline
(3) Questions are presented from most important to least important. \\
\bottomrule
\end{tabular}
}
}
\vspace{-2mm}
\caption{Detailed prompt templates for Reference, AutoCoT, and Metrics generation for Summeval. }
\label{tab:generation_prompt_summeval}
\end{table*}

\definecolor{customcolor1}{RGB}{255, 231, 218}
\definecolor{customcolor2}{RGB}{218, 233, 248}
\definecolor{customcolor3}{RGB}{255, 245, 213}

\begin{table*} [t]
\scriptsize
\centering
\setlength{\tabcolsep}{1.6mm}{
\begin{tabular}{l l >{\raggedright\arraybackslash}m{0.7\textwidth}}
\toprule
\textbf{Template} & \textbf{Value} & \textbf{Prompt} \\
\midrule
\multirow{5}{*}[-7em]{Backbone} &  \multirow{5}{*}[-7em]{-} & 
\begin{tabular}[t]{@{}>{\cellcolor{customcolor1}}p{0.7\textwidth}}
\#\# Instruction \newline
Please act as an impartial judge and evaluate the quality of the response for the next turn in the conversation displayed below on its \textcolor{blue}{[Aspect]}. The response concerns an interesting fact, which will be provided as well. \textcolor{red}{\{reference\_1\_template\}} \textcolor{red}{\{reference\_dialectic\_template\}} \textcolor{red}{\{chain\_of\_thought\_template\}} \\
\end{tabular} \\
& & \  \\
& & 
\begin{tabular}[t]{@{}>{\cellcolor{customcolor2}}p{0.7\textwidth}}
Here are some rules of the evaluation: \newline
1. Your evaluation should consider the \textcolor{blue}{[Aspect]} of the response. \textcolor{blue}{[Criteria]} \newline
2. Be as objective as possible. \newline\newline
\textcolor{red}{\{autocot\_template\}} \\
\end{tabular} \\
& & \ \\
& &
\begin{tabular}[t]{@{}>{\cellcolor{customcolor3}}p{0.7\textwidth}}
\textcolor{red}{\{in\_context\_example\_template\}} \newline
\#\# Conversation History \newline
\textcolor{blue}{[Conversation History]} \newline\newline
\textcolor{red}{\{metrics\_template\}} \newline\newline
\#\# Corresponding Fact \newline
\textcolor{blue}{[Corresponding Fact]} \newline\newline
\textcolor{red}{\{reference\_2\_template\}} \newline\newline
\#\# The Start of Response \newline
\textcolor{blue}{[Response]} \newline
\#\# The End of the Response \\
\end{tabular} \\
\midrule
\multirow{1}{*}{Reference 1} & - & You will also be given a high-quality reference response with the conversation. \\
\midrule
\multirow{1}{*}{Reference 2} & - & \#\# The Start of Reference Response \newline
\textcolor{blue}{[Reference]} \newline
\#\# The End of Reference Response \\
\midrule
\multirow{1}{*}{Reference Dialectic} & - & Please generate your own response for the next turn in the conversation first and take into account your own response to evaluate the quality of the given response. \\
\midrule
\multirow{3}{*}[-1.85em]{Chain-of-Thought} & No CoT & You must directly output your rating of the response on a scale of 1 to \textcolor{red}{\{max\}} without any explanation by strictly following this format: "[[rating]]", for example: "Rating: [[\textcolor{red}{\{max\}}]]". \\
\cmidrule{2-3}
 & Prefix CoT & Begin your evaluation by providing a short explanation. After providing your explanation, you must rate the response on a scale of 1 to \textcolor{red}{\{max\}} by strictly following this format: "[[rating]]", for example: "Rating: [[\textcolor{red}{\{max\}}]]". \\
 \cmidrule{2-3}
 & Suffix CoT & You must rate the response on a scale of 1 to \textcolor{red}{\{max\}} first by strictly following this format: "[[rating]]", for example: "Rating: [[\textcolor{red}{\{max\}}]]". And then provide your explanation. \\
\midrule
\multirow{5}{*}[-1.5em]{Scoring Scale} & 3 &  \textcolor{red}{\{max\}} $ = 3 $ \\
\cmidrule{2-3}
 & 5 & \textcolor{red}{\{max\}} $ = 5 $ \\
\cmidrule{2-3}
 & 10 & \textcolor{red}{\{max\}} $ = 10 $ \\
\cmidrule{2-3}
 & 50 & \textcolor{red}{\{max\}} $ = 50 $ \\
\cmidrule{2-3}
 & 100 & \textcolor{red}{\{max\}} $ = 100 $ \\
\midrule
\multirow{1}{*}{AutoCoT} & - & Evaluation Steps: \newline
\textcolor{blue}{[Autocot]} \\
\midrule
\multirow{1}{*}{In-Context Example} & - & Here are some examples and their corresponding ratings: \newline
\textcolor{red}{\{example\_template\_1\}} \newline
\textcolor{red}{\{example\_template\_2\}} \newline
\textcolor{red}{…} \newline
\textcolor{red}{\{example\_template\_n\}}\newline\newline
Following these examples, evaluate the quality of the response for the next turn in the conversation displayed below on its \textcolor{blue}{[Aspect]}: \\
\midrule
Example & - & \#\# Example \textcolor{blue}{[Number]}:\newline \#\# Conversation History\newline\textcolor{blue}{[Conversation History]}\newline\newline \#\# Corresponding Fact\newline\textcolor{blue}{[Corresponding Fact]}\newline\newline \#\# The Start of the Response \newline\textcolor{blue}{[Response]}\newline \#\# The End of the Response\newline\newline \#\# Rating \newline \textcolor{blue}{[Human Rating]}\\
\midrule
\multirow{1}{*}{Metrics} & - & \#\# Questions about Response \newline
Here are some questions about the response. You can do the evaluation based on thinking about all the questions. \newline
\textcolor{blue}{[Metrics]} \\
\bottomrule
\end{tabular}
}
\vspace{-2mm}
\caption{Detailed evaluation prompt templates for Topical-Chat.}
\label{tab:evaluation_prompt_topical_chat}
\end{table*}
\begin{table*} [t]
\centering
\scriptsize
\resizebox{\linewidth}{!} {
\setlength{\tabcolsep}{1.6mm}{
\begin{tabular}{l >{\raggedright\arraybackslash}m{0.85\textwidth}}
\toprule
\textbf{Template} & \textbf{Prompt} \\
\midrule
\multirow{1}{*}{Reference Generation} & 
Please output the response for the next turn in the conversation. Conversation History: \textcolor{blue}{[Conversation History]} \newline
Response:  \\
\midrule
\multirow{1}{*}{AutoCoT Generation} & 
\#\# Instruction \newline
Please act as an impartial judge and evaluate the quality of the response for the next turn in the conversation on its \textcolor{blue}{[Aspect]} and rate the response on a scale of 1 to \textcolor{red}{\{max\}}.\newline\newline 
Here are some rules of the evaluation: \newline
1. Your evaluation should consider the \textcolor{blue}{[Aspect]} of the response. \textcolor{blue}{[Criteria]} \newline
2. Be as objective as possible. \newline\newline
Please generate the evaluation steps for this task without other explanation.\newline
Evaluation Steps: \\
\midrule
\multirow{1}{*}{Metrics Generation} & 
\#\# Instruction \newline
Please act as an impartial judge and evaluate the quality of the response for the next turn in the conversation displayed below on its \textcolor{blue}{[Aspect]}. Please propose at most three concise questions about whether a potential response is a good response for the next turn in the given conversation on its \textcolor{blue}{[Aspect]}. Another assistant will evaluate the aspect of the output by answering all the questions. \newline

Here are some rules of the evaluation: \newline
(1) Your evaluation should consider the \textcolor{blue}{[Aspect]} of the response. \textcolor{blue}{[Criteria]} \newline
(2) Outputs should NOT contain more/less than what the instruction asks for, as such outputs do NOT precisely execute the instruction. \newline

\#\# Conversation History: \newline
\textcolor{blue}{[Conversation History]} \newline

\#\# Requirements for Your Output: \newline
(1) The questions should **specifically** target the given conversation instead of some general standards, so the questions may revolve around key points of the conversation. \newline
(2) You should directly give the questions without any other words. \newline
(3) Questions are presented from most important to least important. \\
\bottomrule
\end{tabular}
}
}
\vspace{-2mm}
\caption{Detailed prompt templates for Reference, AutoCoT, and Metrics generation for Topical-Chat. }
\label{tab:generation_prompt_topical_chat}
\end{table*}

\definecolor{customcolor1}{RGB}{255, 231, 218}
\definecolor{customcolor2}{RGB}{218, 233, 248}
\definecolor{customcolor3}{RGB}{255, 245, 213}

\begin{table*} [t]
\scriptsize
\centering
\setlength{\tabcolsep}{1.6mm}{
\begin{tabular}{l l >{\raggedright\arraybackslash}m{0.7\textwidth}}
\toprule
\textbf{Template} & \textbf{Value} & \textbf{Prompt} \\
\midrule
\multirow{5}{*}[-7em]{Backbone} &  \multirow{5}{*}[-7em]{-} & 
\begin{tabular}[t]{@{}>{\cellcolor{customcolor1}}p{0.7\textwidth}}
\#\# Instruction \newline
Please act as an impartial judge and evaluate the quality of a natural language sentence generated according to a structured data expression displayed below on its \textcolor{blue}{[Aspect]}. \textcolor{red}{\{reference\_1\_template\}} \textcolor{red}{\{reference\_dialectic\_template\}} \textcolor{red}{\{chain\_of\_thought\_template\}} \\
\end{tabular} \\
& & \  \\
& & 
\begin{tabular}[t]{@{}>{\cellcolor{customcolor2}}p{0.7\textwidth}}
Here are some rules of the evaluation: \newline
1. Your evaluation should consider the \textcolor{blue}{[Aspect]} of the sentence. \textcolor{blue}{[Criteria]} \newline
2. Be as objective as possible. \newline\newline
\textcolor{red}{\{autocot\_template\}} \\
\end{tabular} \\
& & \ \\
& &
\begin{tabular}[t]{@{}>{\cellcolor{customcolor3}}p{0.7\textwidth}}
\textcolor{red}{\{in\_context\_example\_template\}} \newline
\#\# Structured Data Expression \newline
\textcolor{blue}{[Structured Data Expression]} \newline\newline
\textcolor{red}{\{metrics\_template\}} \newline\newline
\textcolor{red}{\{reference\_2\_template\}} \newline\newline
\#\# The Start of the Natural Language Sentence \newline
\textcolor{blue}{[Natural Language Sentence]} \newline
\#\# The End of the Natural Language Sentence \\
\end{tabular} \\
\midrule
\multirow{1}{*}{Reference 1} & - & You will be given the structured data expression, the sentence and a high-quality reference sentence. \\
\midrule
\multirow{1}{*}{Reference 2} & - & \#\# The Start of Reference Sentence \newline
\textcolor{blue}{[Reference]} \newline
\#\# The End of Reference Sentence \\
\midrule
\multirow{1}{*}{Reference Dialectic} & - & Please generate your own sentence according to the given structured data expression first and take into account your own sentence to evaluate the quality of the given sentence. \\
\midrule
\multirow{3}{*}[-1.85em]{Chain-of-Thought} & No CoT & You must directly output your rating of the sentence on a scale of 1 to \textcolor{red}{\{max\}} without any explanation by strictly following this format: "[[rating]]", for example: "Rating: [[\textcolor{red}{\{max\}}]]". \\
\cmidrule{2-3}
 & Prefix CoT & Begin your evaluation by providing a short explanation. After providing your explanation, you must rate the sentence on a scale of 1 to \textcolor{red}{\{max\}} by strictly following this format: "[[rating]]", for example: "Rating: [[\textcolor{red}{\{max\}}]]". \\
 \cmidrule{2-3}
 & Suffix CoT & You must rate the sentence on a scale of 1 to \textcolor{red}{\{max\}} first by strictly following this format: "[[rating]]", for example: "Rating: [[\textcolor{red}{\{max\}}]]". And then provide your explanation. \\
\midrule
\multirow{5}{*}[-1.5em]{Scoring Scale} & 3 &  \textcolor{red}{\{max\}} $ = 3 $ \\
\cmidrule{2-3}
 & 5 & \textcolor{red}{\{max\}} $ = 5 $ \\
\cmidrule{2-3}
 & 10 & \textcolor{red}{\{max\}} $ = 10 $ \\
\cmidrule{2-3}
 & 50 & \textcolor{red}{\{max\}} $ = 50 $ \\
\cmidrule{2-3}
 & 100 & \textcolor{red}{\{max\}} $ = 100 $ \\
\midrule
\multirow{1}{*}{AutoCoT} & - & Evaluation Steps: \newline
\textcolor{blue}{[Autocot]} \\
\midrule
\multirow{1}{*}{In-Context Example} & - & Here are some examples and their corresponding ratings: \newline
\textcolor{red}{\{example\_template\_1\}} \newline
\textcolor{red}{\{example\_template\_2\}} \newline
\textcolor{red}{…} \newline
\textcolor{red}{\{example\_template\_n\}}\newline\newline
Following these examples, evaluate the quality of a natural language sentence generated according to a structured data expression displayed below on its \textcolor{blue}{[Aspect]}: \\
\midrule
Example & - & \#\# Example \textcolor{blue}{[Number]}:\newline \#\# Structured Data Experssion\newline\textcolor{blue}{[Structured Data Experssion]}\newline\newline \#\# The Start of the Natural Language Sentence \newline\textcolor{blue}{[Natural Language Sentence]}\newline \#\# The End of the Natural Language Sentence\newline\newline \#\# Rating \newline \textcolor{blue}{[Human Rating]}\\
\midrule
\multirow{1}{*}{Metrics} & - & \#\# Questions about Sentence \newline
Here are some questions about the sentence. You can do the evaluation based on thinking about all the questions. \newline
\textcolor{blue}{[Metrics]} \\
\bottomrule
\end{tabular}
}
\vspace{-2mm}
\caption{Detailed evaluation prompt templates for SFHOT / SFRES. }
\label{tab:evaluation_prompt_sfhot}
\end{table*}
\begin{table*} [t]
\centering
\scriptsize
\resizebox{\linewidth}{!} {
\setlength{\tabcolsep}{1.6mm}{
\begin{tabular}{l >{\raggedright\arraybackslash}m{0.85\textwidth}}
\toprule
\textbf{Template} & \textbf{Prompt} \\
\midrule
\multirow{1}{*}{Reference Generation} & 
Please generate a natural language sentence generated according to a structured data expression. Expression: \textcolor{blue}{[Expression]} \newline
Sentence:  \\
\midrule
\multirow{1}{*}{AutoCoT Generation} & 
\#\# Instruction \newline
Please act as an impartial judge and evaluate the quality of a natural language sentence generated according to a structured data expression displayed below on its \textcolor{blue}{[Aspect]} and rate the sentence on a scale of 1 to \textcolor{red}{\{max\}}.\newline\newline 
Here are some rules of the evaluation: \newline
1. Your evaluation should consider the \textcolor{blue}{[Aspect]} of the sentence. \textcolor{blue}{[Criteria]} \newline
2. Be as objective as possible. \newline\newline
Please generate the evaluation steps for this task without other explanation.\newline
Evaluation Steps: \\
\midrule
\multirow{1}{*}{Metrics Generation} & 
\#\# Instruction \newline
Please act as an impartial judge and evaluate the quality of a natural language sentence generated according to a structured data expression displayed below on its \textcolor{blue}{[Aspect]}. Please propose at most three concise questions about whether a potential sentence is a good sentence generated according to a given structured data expression on its \textcolor{blue}{[Aspect]}. Another assistant will evaluate the aspects of the sentence by answering all the questions. \newline

Here are some rules of the evaluation: \newline
(1) Your evaluation should consider the \textcolor{blue}{[Aspect]} of the sentence. \textcolor{blue}{[Criteria]} \newline
(2) Outputs should NOT contain more/less than what the instruction asks for, as such outputs do NOT precisely execute the instruction. \newline

\#\# Structured Data Expression: \newline
\textcolor{blue}{[Structured Data Expression]} \newline

\#\# Requirements for Your Output: \newline
(1) The questions should **specifically** target the given structured data expression instead of some general standards, so the questions may revolve around key points of the structured data expressions. \newline
(2) You should directly give the questions without any other words. \newline
(3) Questions are presented from most important to least important. \\
\bottomrule
\end{tabular}
}
}
\vspace{-2mm}
\caption{Detailed prompt templates for Reference, AutoCoT, and Metrics generation for SFHOT / SFRES. }
\label{tab:generation_prompt_sfhot}
\end{table*}

\definecolor{customcolor1}{RGB}{255, 231, 218}
\definecolor{customcolor2}{RGB}{218, 233, 248}
\definecolor{customcolor3}{RGB}{255, 245, 213}

\begin{table*} [t]
\scriptsize
\centering

\setlength{\tabcolsep}{1.6mm}{
\begin{tabular}{l l >{\raggedright\arraybackslash}m{0.7\textwidth}}
\toprule
\textbf{Template} & \textbf{Value} & \textbf{Prompt} \\
\midrule
\multirow{5}{*}[-7em]{Backbone} &  \multirow{5}{*}[-7em]{-} & 
\begin{tabular}[t]{@{}>{\cellcolor{customcolor1}}p{0.7\textwidth}}
\#\# Instruction \newline
Please act as an impartial judge and evaluate the quality of the story generated according to a prompt displayed below on its \textcolor{blue}{[Aspect]}. \textcolor{red}{\{reference\_1\_template\}} \textcolor{red}{\{reference\_dialectic\_template\}} \textcolor{red}{\{chain\_of\_thought\_template\}} \\
\end{tabular} \\
& & \  \\
& & 
\begin{tabular}[t]{@{}>{\cellcolor{customcolor2}}p{0.7\textwidth}}
Here are some rules of the evaluation: \newline
1. Your evaluation should consider the \textcolor{blue}{[Aspect]} of the story. \textcolor{blue}{[Criteria]} \newline
2. Be as objective as possible. \newline\newline
\textcolor{red}{\{autocot\_template\}} \\
\end{tabular} \\
& & \ \\
& &
\begin{tabular}[t]{@{}>{\cellcolor{customcolor3}}p{0.7\textwidth}}
\textcolor{red}{\{in\_context\_example\_template\}} \newline
\#\# Prompt \newline
\textcolor{blue}{[Prompt]} \newline\newline
\textcolor{red}{\{metrics\_template\}} \newline\newline
\textcolor{red}{\{reference\_2\_template\}} \newline\newline
\#\# The Start of the Story \newline
\textcolor{blue}{[Story]} \newline
\#\# The End of the Story \\
\end{tabular} \\
\midrule
\multirow{1}{*}{Reference 1} & - & You will be given the prompt, the generated story and a high-quality reference story. \\
\midrule
\multirow{1}{*}{Reference 2} & - & \#\# The Start of Reference Story \newline
\textcolor{blue}{[Reference]} \newline
\#\# The End of Reference Story \\
\midrule
\multirow{1}{*}{Reference Dialectic} & - & Please generate your own story for the given prompt first and take into account your own story to evaluate the quality of the given story. \\
\midrule
\multirow{3}{*}[-1.85em]{Chain-of-Thought} & No CoT & You must directly output your rating of the story on a scale of 1 to \textcolor{red}{\{max\}} without any explanation by strictly following this format: "[[rating]]", for example: "Rating: [[\textcolor{red}{\{max\}}]]". \\
\cmidrule{2-3}
 & Prefix CoT & Begin your evaluation by providing a short explanation. After providing your explanation, you must rate the story on a scale of 1 to \textcolor{red}{\{max\}} by strictly following this format: "[[rating]]", for example: "Rating: [[\textcolor{red}{\{max\}}]]". \\
 \cmidrule{2-3}
 & Suffix CoT & You must rate the story on a scale of 1 to \textcolor{red}{\{max\}} first by strictly following this format: "[[rating]]", for example: "Rating: [[\textcolor{red}{\{max\}}]]". And then provide your explanation. \\
\midrule
\multirow{5}{*}[-1.5em]{Scoring Scale} & 3 &  \textcolor{red}{\{max\}} $ = 3 $ \\
\cmidrule{2-3}
 & 5 & \textcolor{red}{\{max\}} $ = 5 $ \\
\cmidrule{2-3}
 & 10 & \textcolor{red}{\{max\}} $ = 10 $ \\
\cmidrule{2-3}
 & 50 & \textcolor{red}{\{max\}} $ = 50 $ \\
\cmidrule{2-3}
 & 100 & \textcolor{red}{\{max\}} $ = 100 $ \\
\midrule
\multirow{1}{*}{AutoCoT} & - & Evaluation Steps: \newline
\textcolor{blue}{[Autocot]} \\
\midrule
\multirow{1}{*}{In-Context Example} & - & Here are some examples and their corresponding ratings: \newline
\textcolor{red}{\{example\_template\_1\}} \newline
\textcolor{red}{\{example\_template\_2\}} \newline
\textcolor{red}{…} \newline
\textcolor{red}{\{example\_template\_n\}}\newline\newline
Following these examples, evaluate the story generated according to a prompt displayed below on its \textcolor{blue}{[Aspect]}: \\
\midrule
Example & - & \#\# Example \textcolor{blue}{[Number]}:\newline \#\# Prompt\newline\textcolor{blue}{[Prompt]}\newline\newline \#\# The Start of the Story \newline\textcolor{blue}{[Story]}\newline \#\# The End of the Story\newline\newline \#\# Rating \newline \textcolor{blue}{[Human Rating]}\\
\midrule
\multirow{1}{*}{Metrics} & - & \#\# Questions about Story \newline
Here are some questions about the story. You can do the evaluation based on thinking about all the questions. \newline
\textcolor{blue}{[Metrics]} \\
\bottomrule
\end{tabular}
}

\vspace{-2mm}
\caption{Detailed evaluation prompt templates for HANNA. }
\label{tab:evaluation_prompt_hanna}
\end{table*}
\begin{table*} [t]
\centering
\scriptsize
\resizebox{\linewidth}{!} {
\setlength{\tabcolsep}{1.6mm}{
\begin{tabular}{l >{\raggedright\arraybackslash}m{0.85\textwidth}}
\toprule
\textbf{Template} & \textbf{Prompt} \\
\midrule
\multirow{1}{*}{Reference Generation} & 
Please generate a story according to the given prompt: \textcolor{blue}{[Prompt]} \newline
Story:  \\
\midrule
\multirow{1}{*}{AutoCoT Generation} & 
\#\# Instruction \newline
Please act as an impartial judge and evaluate the quality of the story generated according to a prompt displayed below on its \textcolor{blue}{[Aspect]} and rate the story on a scale of 1 to \textcolor{red}{\{max\}}.\newline\newline 
Here are some rules of the evaluation: \newline
1. Your evaluation should consider the \textcolor{blue}{[Aspect]} of the story. \textcolor{blue}{[Criteria]} \newline
2. Be as objective as possible. \newline\newline
Please generate the evaluation steps for this task without other explanation.\newline
Evaluation Steps: \\
\midrule
\multirow{1}{*}{Metrics Generation} & 
\#\# Instruction \newline
Please act as an impartial judge and evaluate the quality of the story generated according to a prompt displayed below on its \textcolor{blue}{[Aspect]}. Please propose at most three concise questions about whether a potential story is a good story according to a given prompt on its \textcolor{blue}{[Aspect]}. Another assistant will evaluate the aspect of the story by answering all the questions. \newline

Here are some rules of the evaluation: \newline
(1) Your evaluation should consider the \textcolor{blue}{[Aspect]} of the story. \textcolor{blue}{[Criteria]} \newline
(2) Outputs should NOT contain more/less than what the instruction asks for, as such outputs do NOT precisely execute the instruction. \newline

\#\# Prompt: \newline
\textcolor{blue}{[Prompt]} \newline

\#\# Requirements for Your Output: \newline
(1) The questions should **specifically** target the given prompt instead of some general standards, so the questions may revolve around key points of the prompt. \newline
(2) You should directly give the questions without any other words. \newline
(3) Questions are presented from most important to least important. \\
\bottomrule
\end{tabular}
}
}
\vspace{-2mm}
\caption{Detailed prompt templates for Reference, AutoCoT, and Metrics generation for HANNA. }
\label{tab:generation_prompt_hanna}
\end{table*}

\begin{table*} [!t]
\centering
\resizebox{\linewidth}{!} {
\begin{tabular}{cccc|c|c|c|c|c|c|c|c|c}
\toprule
\multirow{2}{*}{\textbf{Model}} &
\multirow{2}{*}{\textbf{Dataset}} &
\multirow{2}{*}{\textbf{Aspect}} &
\textbf{During} &
\multicolumn{8}{c|}{\textbf{Prompting Strategy}} &
\textbf{Spearman} \\
\cmidrule(lr){5-12}
& &  &
\textbf{HPSS} &
 Scale & Example & Criteria & Reference & CoT & AutoCoT & Metrics & Order & 
 \multicolumn{1}{c}{\textbf{Correlation}}\\
\midrule
\multirow{6}{*}{\textbf{GPT-4o-mini}} & \multirow{3}{*}{SFHOT} & \multirow{3}{*}{Naturalness} &
Origin & 5 & 0 & Human & \XSolidBrush & Prefix & \XSolidBrush & \XSolidBrush & \textbf{TD $\rightarrow$ ER $\rightarrow$ IC} & 0.366 \\
 & & & Best & 100 & 0 & Human & \XSolidBrush & \XSolidBrush & \XSolidBrush & \XSolidBrush & \textbf{TD $\rightarrow$ ER $\rightarrow$ IC} & 0.437 \textcolor{red}{\small (+19.4\%)} \\
 & & & Worst & 5 & 0 & Human & Dialectic & Prefix &
\XSolidBrush & \XSolidBrush & \textbf{TD $\rightarrow$ ER $\rightarrow$ IC} & 0.104 \textcolor{blue}{\small (-71.6\%)} \\
\cmidrule(lr){2-13}
 & \multirow{3}{*}{HANNA} & \multirow{3}{*}{Complexity} &
Origin & 5 & 0 & Human & \XSolidBrush & Prefix & \XSolidBrush & \XSolidBrush & \textbf{TD $\rightarrow$ ER $\rightarrow$ IC} & 0.458\\
 & & & Best & 100 & 10 & Self-Generated & \XSolidBrush & \XSolidBrush &
\CheckmarkBold & \XSolidBrush & 
\textbf{IC $\rightarrow$ ER $\rightarrow$ TD} & 0.617 \textcolor{red}{\small (+34.7\%)} \\
 & & & Worst & 5 & 0 & Human & Dialectic & Prefix &
\XSolidBrush & \XSolidBrush & \textbf{TD $\rightarrow$ ER $\rightarrow$ IC} & -0.046 \textcolor{blue}{\small (-110.0\%)}\\
\cmidrule(lr){1-13}
\multirow{6}{*}{\textbf{Qwen2.5-14B}} & \multirow{3}{*}{SFRES} & \multirow{3}{*}{Informativeness} &
Origin & 5 & 0 & Human & \XSolidBrush & Prefix & \XSolidBrush & \XSolidBrush & \textbf{TD $\rightarrow$ ER $\rightarrow$ IC} & 0.169\\
 & & & Best & 50 & 3 & \XSolidBrush & 
 Dialectic & Prefix &
\XSolidBrush & \XSolidBrush & \textbf{TD $\rightarrow$ IC $\rightarrow$ ER} & 0.381 \textcolor{red}{\small (+125.4\%)} \\
 & & & Worst & 5 & 5 & \XSolidBrush & \XSolidBrush & Prefix &
\XSolidBrush & \XSolidBrush & \textbf{ER $\rightarrow$ IC $\rightarrow$ TD} & 0.088 \textcolor{blue}{\small (-47.9\%)}\\
\cmidrule(lr){2-13}
 & \multirow{3}{*}{Topical-Chat} & \multirow{3}{*}{Coherence} &
Origin & 3 & 0 & Human & \XSolidBrush & Prefix & \XSolidBrush & \XSolidBrush & \textbf{TD $\rightarrow$ ER $\rightarrow$ IC} & 0.520\\
 & & & Best & 10 & 0 & \XSolidBrush & Self-Generated & Suffix &
\CheckmarkBold & 
\CheckmarkBold & \textbf{ER $\rightarrow$ IC $\rightarrow$ TD} & 0.651 \textcolor{red}{\small (+25.2\%)} \\
 & & & Worst & 10 & 5 & \XSolidBrush & Dialectic & \XSolidBrush &
\CheckmarkBold & \XSolidBrush & \textbf{ER $\rightarrow$ TD $\rightarrow$ IC} & 0.456 \textcolor{blue}{\small (-12.3\%)}\\
\bottomrule
\end{tabular}
}
\vspace{-2mm}
\caption{
Illustration of some prompting strategies explored by HPSS and their performance on the validation dataset. 
We present both the best-performing and worst-performing prompting strategies for specific tasks during HPSS.
} 
\label{tab:case_study}
\end{table*}
\section{Case Study}
\label{app:case}
\begin{figure*}[h]
\scriptsize
    \centering
    \includegraphics[width=0.95\textwidth]{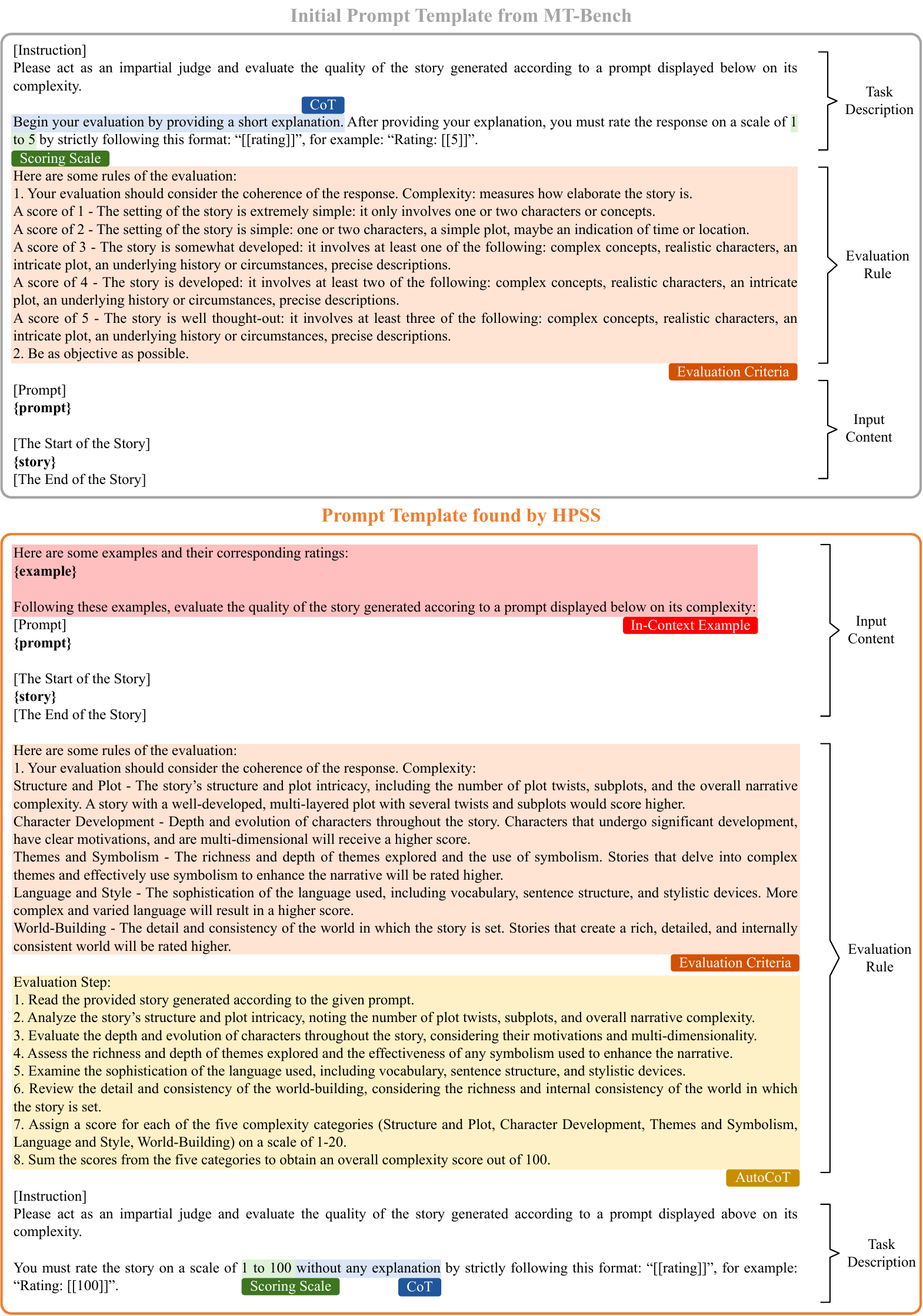}
    \vspace{-2mm}
    \caption{The original evaluation prompt for the aspect \textit{Complexity} in HANNA from MT-Bench and the corresponding evaluation prompt found by HPSS for GPT-4o-mini evaluator.
    Factors with values other than "None" in the evaluation prompts are highlighted. 
    The three main components of the evaluation prompt are annotated on the right side, e.g., Task Description (\textbf{TD}), Evaluation Rule (\textbf{ER}), and Input Content (\textbf{IC}). 
    The placement order of these three components is also considered a factor in our optimization.}
    \label{fig:case_1}
\end{figure*}

\begin{figure*}[h]
\scriptsize
    \centering
    \includegraphics[width=0.95\textwidth]{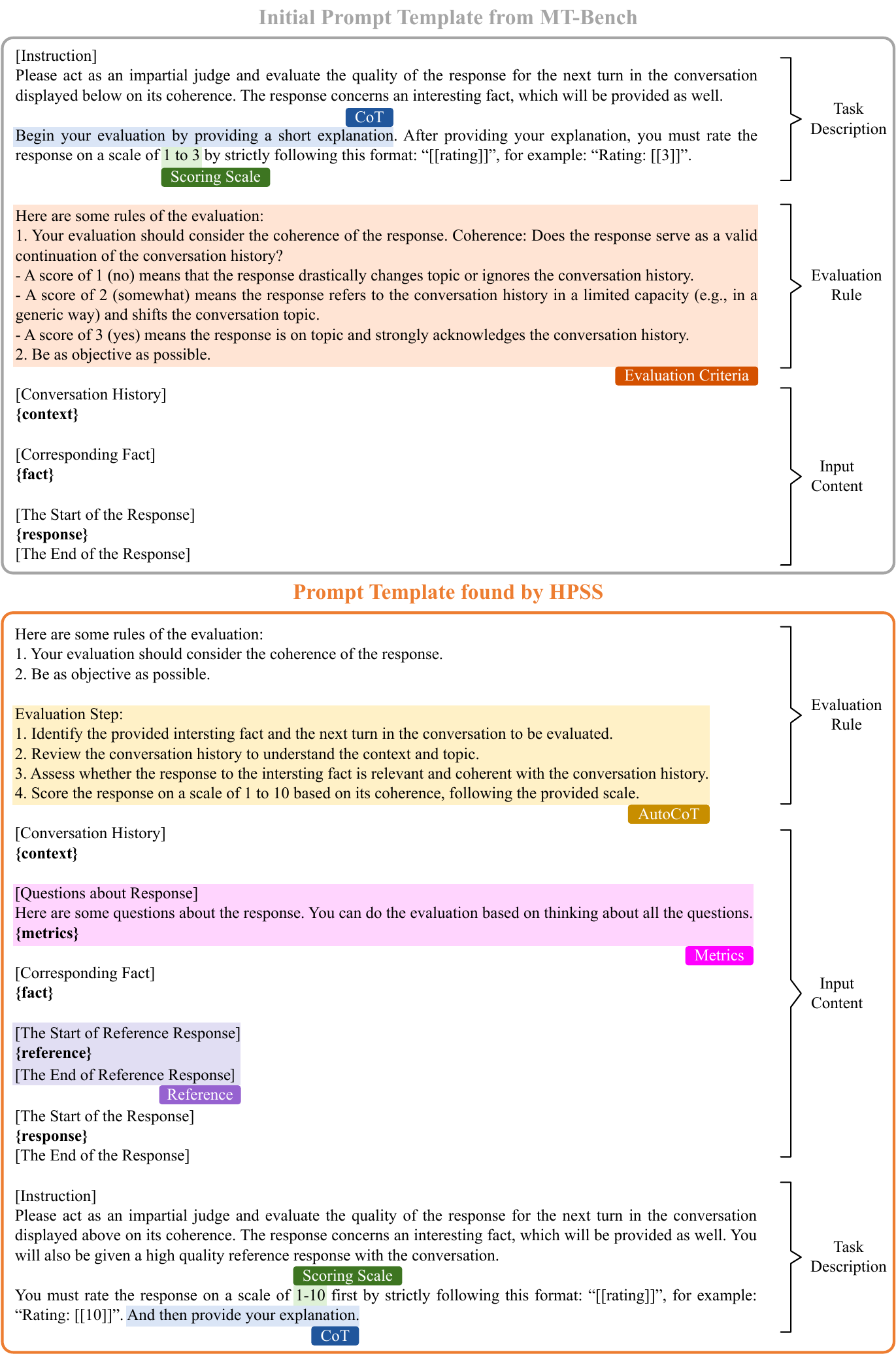}
    \vspace{-2mm}
    \caption{The original evaluation prompt for the aspect \textit{Coherence} in Topical-Chat from MT-Bench and the corresponding evaluation prompt found by HPSS for Qwen2.5-14B-Instruct evaluator.}
    \label{fig:case_2}
\end{figure*}

In Table \ref{tab:case_study}, we present some cases of prompting strategies explored by HPSS for specific tasks. 
We find that there is a significant performance gap between the best-performing and worst-performing prompting strategies, 
emphasizing the sensitivity of LLM evaluators to the prompting strategy. 
Figure \ref{fig:case_1} and \ref{fig:case_2} illustrate the original evaluation prompts from MT-Bench alongside the optimized one found by HPSS, specifically focusing on the aspect \textit{Complexity} in HANNA and the aspect \textit{Coherence} in Topical-Chat.
The prompting strategies of these prompts are shown in the "origin" and "best" rows in Table \ref{tab:case_study} for their respective datasets and aspects.
We observe that some of the prompting strategies found by HPSS include values that are rarely considered in human-designed evaluation prompts. 
For instance, for the aspect \textit{Complexity} in HANNA, the prompting strategy found by HPSS places the input content (\textbf{IC}) first and the task description (\textbf{TD}) last.
For the aspect \textit{Coherence} in Topical-Chat, evaluation criteria are not used in HPSS. 
These results demonstrate the limitations of manual prompt design and underscore the importance of automatic prompting strategy optimization.

\begin{figure*}[h]
\scriptsize
    \centering
    \includegraphics[width=0.95\textwidth]{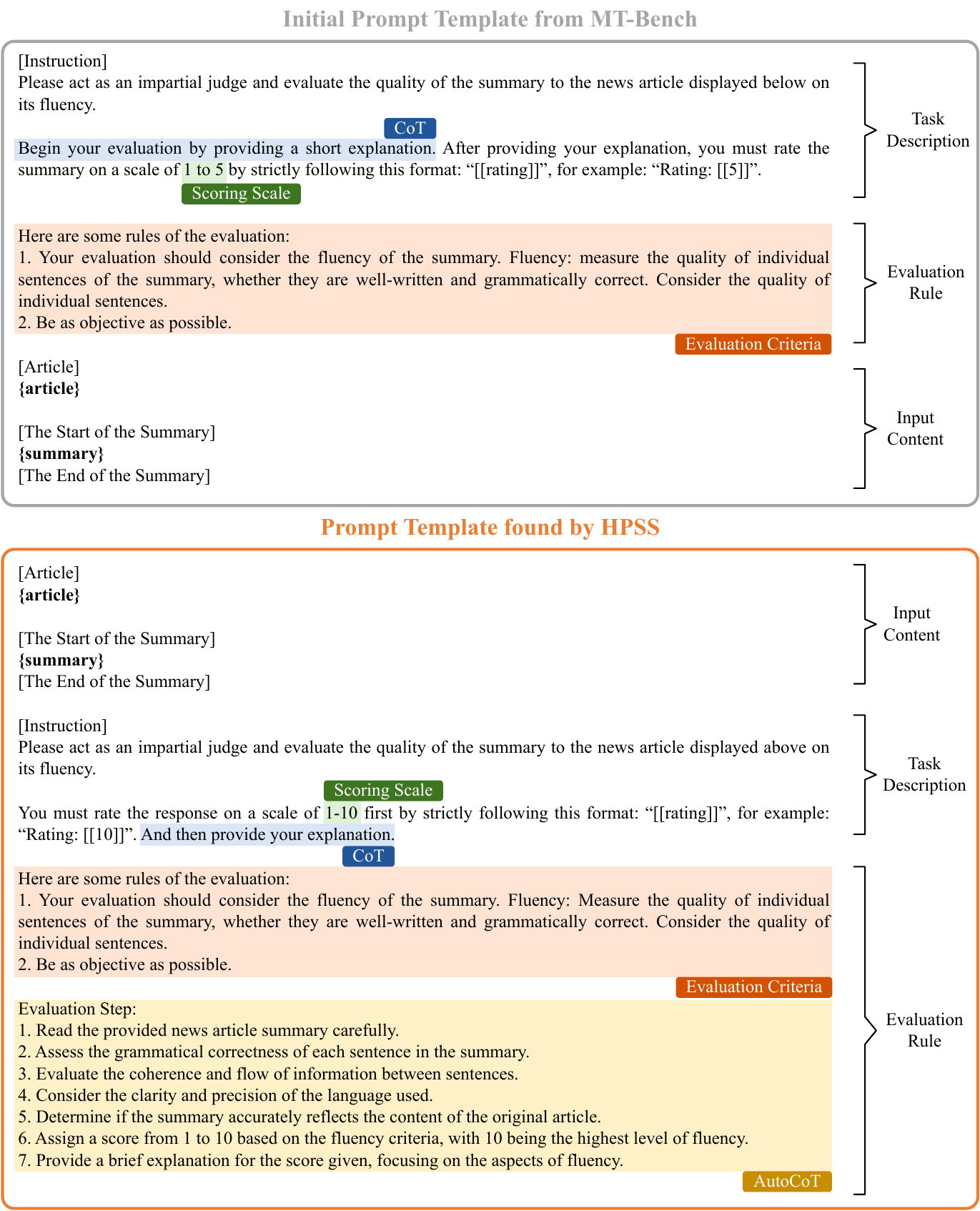}
    \vspace{-2mm}
    \caption{The original evaluation prompt for the aspect \textit{Fluency} in Summeval from MT-Bench and the corresponding evaluation prompt found by HPSS for Qwen2.5-14B-Instruct evaluator.}
    \label{fig:case_3}
\end{figure*}

\begin{figure*}[h]
\scriptsize
    \centering
    \includegraphics[width=0.95\textwidth]{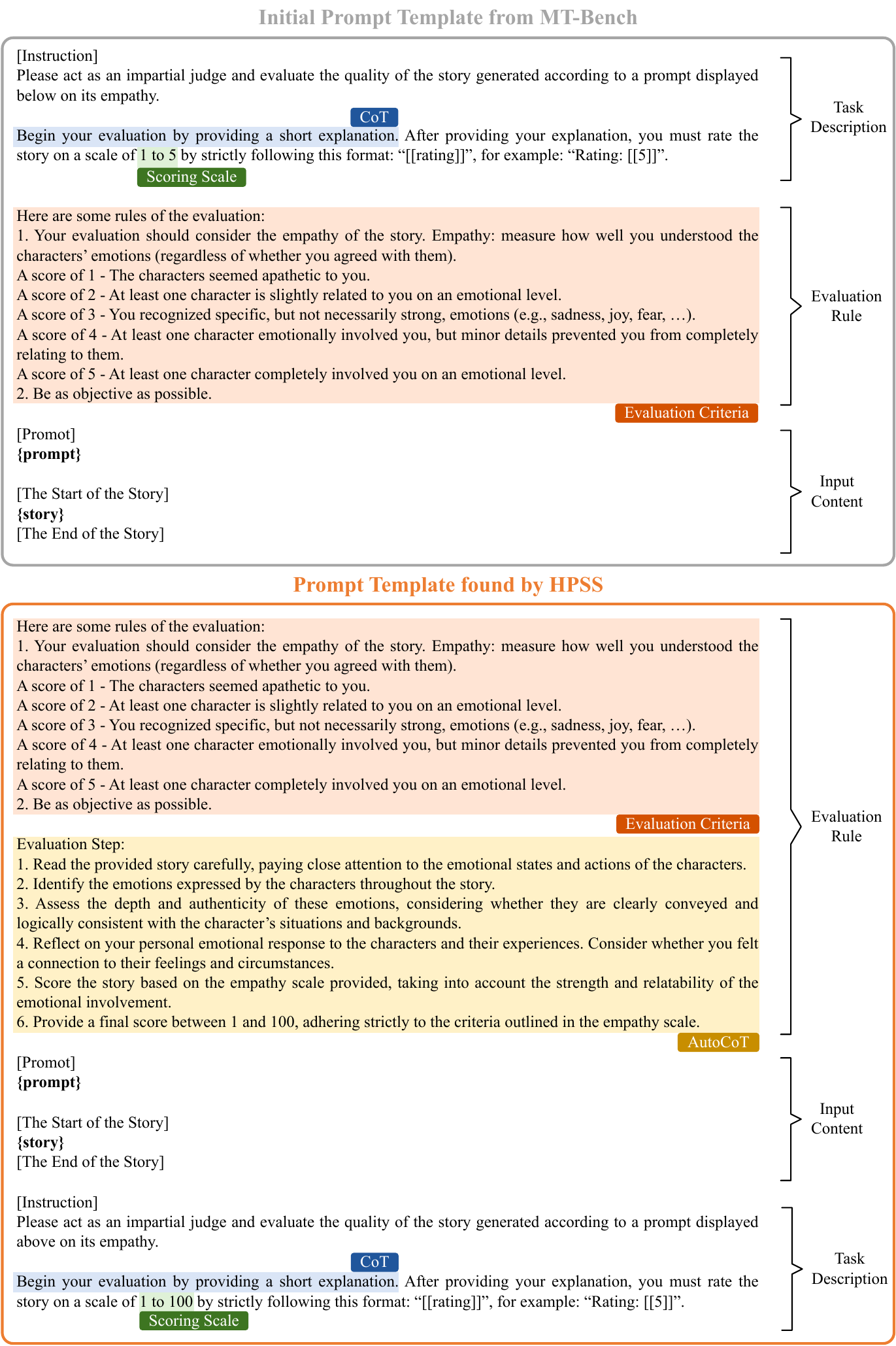}
    \vspace{-2mm}
    \caption{The original evaluation prompt for the aspect \textit{Empathy} in HANNA from MT-Bench and the corresponding evaluation prompt found by HPSS for Qwen2.5-14B-Instruct evaluator.}
    \label{fig:case_4}
\end{figure*}

Furthermore, to intuitively show the reason why HPSS achieves better evaluation performance than human-designed LLM evaluators, we provide two judgment generation cases for different LLM evaluators in Table \ref{tab:judgement_case_1} and \ref{tab:judgement_case_2}. 
The corresponding evaluation prompts are shown in Figure \ref{fig:case_3} and \ref{fig:case_4}.
In the first case from the text summarization task presented in Table \ref{tab:judgement_case_1}, HPSS provides a balanced assessment by analyzing both strengths and weaknesses of the summary, with emphasis on overall fluency. 
In contrast, MT-Bench and CloserLook + ICL overemphasize minor issues like typos while overlooking its advantages in overall fluency, and exhibit some hallucinations in judgment.
In the second case from the story generation task presented in Table \ref{tab:judgement_case_2},
HPSS conducts a systematic analysis of the story and correctly identifies both key plots that effectively convey emotions and overly idealized plots that weaken emotional delivery. 
In contrast, MT-Bench and CloserLook + ICL each overlook one of these two important points, resulting in less accurate evaluations. 
These observations demonstrate that HPSS improves the ability of LLM evaluators to conduct comprehensive evaluations
and achieve a balanced assessment of strengths and weaknesses within the input sample.

\begin{table*}[!ht]
\scriptsize
    \centering
    \begin{tabular}{p{55pt}p{365pt}}
    \toprule
    Article & Chelsea have made an offer for FC Tokyo's 22-year-old forward Yoshinori Muto, according to club president Naoki Ogane. \newline
    The Japan international, who has played for the J-League side since 2013, will join Chelsea's Dutch partner club Vitesse Arnhem on loan next season if he completes a move to Stamford Bridge this summer. \newline
    Ogane claims that Chelsea's interest in Muto is not connected to the £200million sponsorship deal they signed with Japanese company Yokohama Rubber in February. \newline
    \textbf{FC Tokyo forward Yoshinori Muto (centre) brings the ball forward against Albirex Niigata in March.}\newline
    \textbf{FC Tokyo president Naoki Ogane claims that Chelsea have made a bid for Japan international Muto.}\newline
    \textbf{Muto tussles with Yuji Nakazawa of Yokohama F.Marinos during a J-League clash last month.}\newline\newline
    \textbf{YOSHINORI MUTO FACTFILE} \newline
    \textbf{Age}: 22 \newline
    \textbf{Club}: FC Tokyo \newline
    \textbf{Appearances}: 37 \newline
    \textbf{Goals}: 16 \newline
    \textbf{International caps (Japan)}: 11 \newline
    \textbf{International goals}: 1 \newline
    \textbf{Did you know?} Muto graduated from Keio University in Tokyo with an economics degree two weeks ago. \newline\newline
    Speaking to Sports Nippon, Ogane said: 'It is true that Chelsea sent us an offer for Muto. 'It is a formal offer with conditions. They want to acquire him in the summer.' \newline
    Muto, who only graduated from Keio University a fortnight ago after completing an economics degree, would be the first Japanese player to represent Chelsea if he moves to west London. He has earned 11 caps for his country after signing his first professional contract in 2014, scoring once for the Samurai Blue. \newline
    A £4million deal for the youngster has been mooted, but Muto admits that he isn't sure if he will join the Premier League title chasers despite being pleased with their bid. \newline
    He said: 'I have not decided yet at all. It is an honour for me to receive this offer from a great club.' \newline
    Muto scored 13 times in his debut season with FC Tokyo and was named in the J-League's best XI. \newline
    \textbf{Muto admits it is an 'honour' to receive an offer from Chelsea although he has not yet decided to join the club.} \newline
    \textbf{Muto, pictured in action against Ventforet Kofu has scored three goals in four games so far this season.} \newline
    \textbf{The 22-year-old has a shot at goal during Japan's Asian Cup match against Palestine in January this year.}\newline 
    He has continued his fine form during the current campaign, helping his club to third place in the division with three goals in four games. \newline
    Yokohama Rubber - one of the world's largest tyre manufacturers - will become Chelsea's official shirt sponsors from the start of the 2015-16 season. \newline
    The initial five-year deal is the biggest in the club's history, with the Blues now considering a two-week pre-season tour of Japan this summer. \\
    \midrule
    Summary & Chelsea have made an offer for fc tokyo's 22-year-old forward yoshinori muto. 
    The japan international will join chelsea's dutch partner club vitesse arnhem on loan next season if he completes a move to stamford bridge this summer. 
    Muto claims chelsea's interest in muto is not connected to the \# 200million sponsorship deal they signed with japanese company yokohama rubber in february.  \\
    \midrule
    Evaluation Aspect  & Fluency \\
    \midrule
    Human Score (1-5) & 5 \\
    \midrule
    \midrule
    MT-Bench (1-5)  & The summary contains several issues related to fluency, \textcolor{red}{primarily due to grammatical errors}, inconsistencies in capitalization, and the inclusion of a symbol (\#) that does not belong. \textcolor{red}{The sentences are generally clear but lack the smoothness and refinement that one would expect in a quality summary.} The use of lower-case letters for "fc tokyo" and "stamford bridge" is incorrect, and the number symbol (\#) before "200million" is out of place. There are also minor issues with sentence structure and flow, such as the abrupt transition from the offer to the loan arrangement without proper context.\newline\newline
    Rating: [[2]] \\
    \midrule
    CloserLook + ICL (1-5) & Analysis: The summary contains several grammatical errors and lacks coherence. Sentence fragments and improper punctuation are present, making the summary difficult to read. \textcolor{red}{For example, "The japan international will join chelsea's vitesse arnhem on loan next season if he completes a move to stamford bridge this summer" is not a complete sentence and lacks clarity.} Additionally, the summary includes a hashtag (\# 200million) instead of writing out the number, which is inconsistent with standard writing practices. The summary also omits some important details from the original article, such as Muto's achievements and his indecision about the move.\newline\newline
    Rating: 1  \\
    \midrule 
    HPSS (1-10)  & Rating: [[8]]\newline\newline
    Explanation:\newline
    The summary is generally well-written and grammatically correct. The sentences flow well and are coherent, providing a clear and concise overview of the key points from the article. However, there are a few areas where the fluency could be improved:\newline\newline
    1. The summary starts with a run-on sentence that combines two separate ideas from the article: "Chelsea have made an offer for fc tokyo's 22-year-old forward yoshinori muto. The japan international will join chelsea's vitesse arnhem on loan next season if he completes a move to stamford bridge this summer." This could be split into two sentences for better readability: "Chelsea have made an offer for FC Tokyo's 22-year-old forward Yoshinori Muto. If he completes a move to Stamford Bridge this summer, the Japan international will join Chelsea's Vitesse Arnhem on loan next season."\newline\newline
    2. The use of "\# 200million" instead of "200 million" is inconsistent with the currency used in the original article and can be confusing for readers.\newline\newline
    3. While the summary is concise, it could include more details to better reflect the depth of the original article. For instance, mentioning Muto's form this season or his academic background could enhance the summary without significantly increasing its length.\newline\newline
    Overall, the summary is clear and accurately reflects the main points of the article, but a few minor adjustments could improve its fluency and completeness. \\
    \bottomrule
    \end{tabular}
    \vspace{-2mm}
    \caption{A judgment generation case of Qwen2.5-14B-Instruct evaluator using the prompting strategy from MT-Bench, CloserLook + ICL, and HPSS for the aspect \textit{Fluency} in Summeval. Factual errors in the judgments are highlighted in \textcolor{red}{red}.}
    \label{tab:judgement_case_1}
\end{table*}
\begin{table*}[!ht]
\scriptsize
    \centering
    \begin{tabular}{p{55pt}p{365pt}}
    \toprule
    Prompt & One normal day you see a mouse pointer move across your field of vision, right click, and delete an object. \\
    \midrule
    Story & It was a long tiring day as usual. I leaned on my sofa throwing my bag on to the near by table. Slowly dozed off to sleep. My imaginations rushed in, I could literally feel my dream and it was of none other than an episode which caught my attention today. Me and one of my colleague were having lunch and a small boy standing outside the canteen was cleaning the window panes, he was hardly 10. Full greased head to foot, filthy clothes, torn shoes but with a cherry red lips and a droopy eyes filled with all innocence he could gather within it. It was raining and he was shivering after a while. He sat loopy near the window base facing me. I was having coffee then. As we started staring at each other I felt difficulty in drinking my coffee. It was nothing but me sitting comfortably inside the cafe warm and the boy now half drenched killed me within. I could n't read any thing from his face that he is not liking what he is going through. I called him and offered him a tea. Gladly he accepted it! I asked dont you feel cold, he told no I feel close to nature. I was so taken by the love and spirit he had within and his love for nature, the smallest things which I could n't realize which he made me feel was aw inspiring. I wish I had a mouse with me to take of the poverty in his life with a click, and fill his childhood with all the beautiful things he deserve. And soon i realized this was a beautiful dream, a dream that i would love it to come true. Daffodils  \\
    \midrule
    Evaluation Aspect  & Empathy \\
    \midrule
    Human Score (1-5) & 3.67 \\
    \midrule
    \midrule
    MT-Bench (1-5)  & The story does a remarkable job of evoking empathy by deeply immersing the reader in the narrator's emotional journey. The narrator's guilt and compassion are palpable as they reflect on the stark contrast between their comfortable existence and the boy's challenging circumstances. The emotional resonance is further heightened by the narrator's wish to alleviate the boy's poverty with a simple click, symbolizing a desire to change the world through a small, symbolic action. The dream sequence provides a poignant backdrop that amplifies the narrator's sentiments.\newline\newline Rating: [[5]] \\
    \midrule
    CloserLook + ICL (1-5) & Analysis: The story does not directly address the prompt about a mouse pointer deleting an object. Instead, it focuses on a dream about a young boy cleaning window panes in the rain, highlighting the character's emotional response to the boy's innocence and resilience. The narrator feels a deep sense of empathy and guilt, wishing they could alleviate the boy's poverty. While the story does evoke strong emotions, it does not connect these emotions to the prompt's scenario involving a mouse pointer.\newline\newline
    Rating: 3  \\
    \midrule 
    HPSS (1-100)  & The story delves into the narrator's emotional response to a young boy cleaning window panes outside a canteen during a rainy day. The narrator's empathy is palpable as they describe the stark contrast between their comfortable situation inside the cafe and the boy's difficult circumstances. The emotions conveyed include compassion, guilt, and a strong desire to alleviate the boy's poverty. The narrator's emotional journey from discomfort to connection is well-articulated and relatable. However, the narrative occasionally veers into a dream sequence that includes the wish for a magical solution to poverty, which slightly detracts from the realism and depth of the emotional engagement. Despite this, the core emotional experience is effectively conveyed and deeply felt.\newline\newline
    Rating: [[75]] \\
    \bottomrule
    \end{tabular}
    \vspace{-2mm}
    \caption{A judgment generation case of Qwen2.5-14B-Instruct evaluator using the prompting strategy from MT-Bench, CloserLook + ICL, and HPSS for the aspect \textit{Empathy} in HANNA.}
    \label{tab:judgement_case_2}
\end{table*}

\end{document}